%% file: ms.tex
\pdfoutput=1
\documentclass[10pt,twocolumn,letterpaper]{article}

\usepackage[pagenumbers]{cvpr} 

\usepackage{fancyhdr}
\usepackage{graphicx}
\usepackage{amsmath}
\usepackage{amssymb}
\usepackage{booktabs}
\usepackage{siunitx}
\usepackage{ amssymb }
\usepackage{makecell}
\usepackage{color, colortbl}
\usepackage[accsupp]{axessibility}  

\definecolor{Gray}{gray}{0.9}

\DeclareMathOperator{\LM}{\mathcal{L}}

\DeclareMathOperator{\NM}{\mathcal{N}}

\makeatletter
\newcommand*\bigcdot{\mathpalette\bigcdot@{.5}}
\newcommand*\bigcdot@[2]{\mathbin{\vcenter{\hbox{\scalebox{#2}{$\m@th#1\bullet$}}}}}
\makeatother

\ExplSyntaxOn
\NewDocumentCommand{\longdash}{ O{2} }
 {
  --\prg_replicate:nn { #1 - 1 } { \negthinspace -- }
 }
\ExplSyntaxOff

\usepackage[pagebackref,breaklinks,colorlinks,hypertexnames=false]{hyperref}

\usepackage[capitalize]{cleveref}
\crefname{section}{Sec.}{Secs.}
\Crefname{section}{Section}{Sections}
\Crefname{table}{Table}{Tables}
\crefname{table}{Tab.}{Tabs.}

\def\cvprPaperID{5588} 
\def\confName{CVPR}
\def\confYear{2022}

\begin{document}
\pagestyle{fancy}
\setlength{\headheight}{6.0pt}
\setlength{\headsep}{4mm}
\topmargin -0.625in
\addtolength{\headsep}{0.25in}
\fancyhead{}
\fancyhead[L]{Accepted for publication at CVPR 2023}
\setlength{\footskip}{50pt}

\title{Best of Both Worlds: Multimodal Contrastive Learning with Tabular and Imaging Data}

\author{Paul Hager\textsuperscript{1,2} \qquad
    Martin J. Menten\textsuperscript{1,2,3} \qquad Daniel Rueckert\textsuperscript{1,2,3} \\
    \textsuperscript{1}Technical University of Munich, \textsuperscript{2}Klinikum Rechts der Isar, \textsuperscript{3}Imperial College London \\
    {\tt\small \{paul.hager, martin.menten, daniel.rueckert\}@tum.de}
}
\maketitle
\thispagestyle{fancy}

\begin{abstract}
    
    Medical datasets and especially biobanks, often contain extensive tabular data with rich clinical information in addition to images.
    In practice, clinicians typically have less data, both in terms of diversity and scale, but still wish to deploy deep learning solutions.
    Combined with increasing medical dataset sizes and expensive annotation costs, the necessity for unsupervised methods that can pretrain multimodally and predict unimodally has risen.
    
    To address these needs, we propose the first self-supervised contrastive learning framework that takes advantage of images and tabular data to train unimodal encoders.
    Our solution combines SimCLR and SCARF, two leading contrastive learning strategies, and is simple and effective.
    In our experiments, we demonstrate the strength of our framework by predicting risks of myocardial infarction and coronary artery disease (CAD) using cardiac MR images and 120 clinical features from 40,000 UK Biobank subjects.
    Furthermore, we show the generalizability of our approach to natural images using the DVM car advertisement dataset.

    We take advantage of the high interpretability of tabular data and through attribution and ablation experiments find that morphometric tabular features, describing size and shape, have outsized importance during the contrastive learning process and improve the quality of the learned embeddings.
    Finally, we introduce a novel form of supervised contrastive learning, label as a feature (LaaF), by appending the ground truth label as a tabular feature during multimodal pretraining, outperforming all supervised contrastive baselines.\footnote{\url{https://github.com/paulhager/MMCL-Tabular-Imaging}}
    
    
\end{abstract}

\section{Introduction}
\label{sec:intro}

\begin{figure*}
  \centering
  \includegraphics[width=\linewidth]{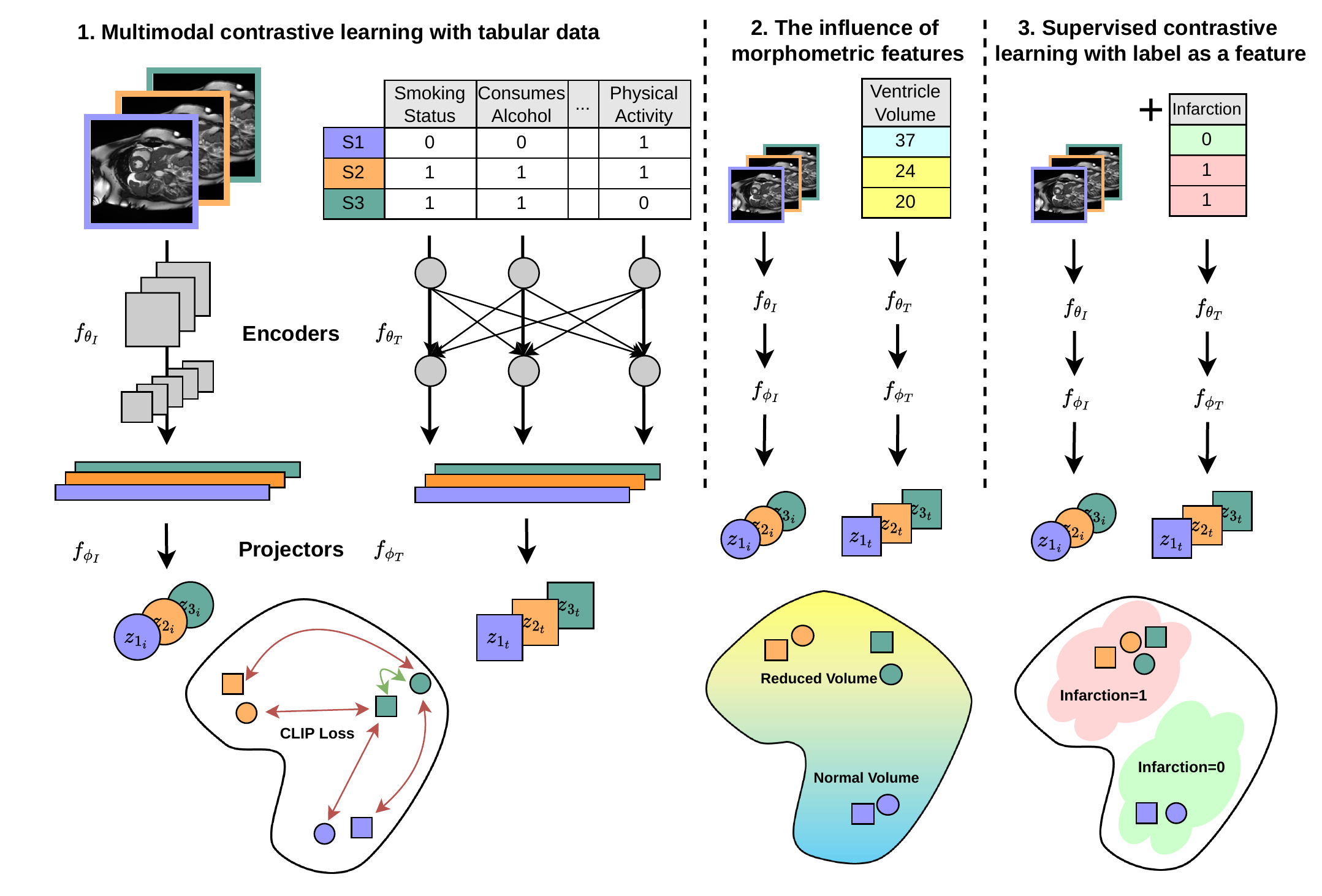}
  \caption{We combine imaging and tabular data in a contrastive learning framework. We observe that morphometric features, describing shape and size, are of outsized importance in multimodal contrastive training and their inclusion boosts downstream task performance. By simply adding the label as a tabular feature we introduce a novel form of supervised contrastive learning that outperforms all other supervised contrastive strategies.}
  \label{fig:graphicalabstract}
\end{figure*}

Modern medical datasets are increasingly multimodal, often incorporating both imaging and tabular data. 
Images can be acquired by computed tomography, ultrasound, or magnetic resonance scanners, while tabular data commonly originates from laboratory tests, medical history and patient lifestyle questionnaires.
Clinicians have the responsibility to combine and interpret this tabular and imaging data to diagnose, treat, and monitor patients.
For example, cardiologists may ask about a patients' family history and record their weight, cholesterol levels, and blood pressure to better inform diagnoses when examining images of their heart.

Beyond diagnostics, multimodal data is also crucial to advance the understanding of diseases motivating the creation of biobanks.
Going far beyond the scale of typical datasets in hospitals, biobanks pool vast amount of information from large populations. 
Multimodal biobanks include the German National Cohort \cite{noauthor_german_2014} with 200,000 subjects, Lifelines\cite{sijtsma_cohort_2022} with 167,000 subjects, and the UK Biobank\cite{sudlow_uk_2015} with 500,000 subjects. 
The UK Biobank includes thousands of data fields from patient questionnaires, laboratory tests, and medical examinations, in addition to imaging and genotyping information.
Biobanks have already proven useful in the training of machine learning models to predict many diseases such as anaemia\cite{mitani_detection_2020}, early brain aging\cite{jonsson_brain_2019} and cardiovascular disease\cite{alaa_cardiovascular_2019, rim_deep-learning-based_2021}.

There is a substantial interest in deploying algorithms that have been developed using these large-scale population studies in clinical practice. 
However, acquiring the same quality of data, both in terms of diversity of modalities and number of features, is not feasible in a busy clinical workflow\cite{dugdale_time_1999}. 
Furthermore, low disease frequencies make supervised solutions hard to train.
Consequently, there is a clear need for unsupervised strategies that can learn from biobank scale datasets and be applied in the clinic where considerably less data, in size and dimension, is available.

\paragraph{Our contribution}
To address these needs, we propose the first contrastive framework that utilizes imaging and tabular data, shown in figure \ref{fig:graphicalabstract}. 
Our framework is based on SimCLR\cite{chen_simple_2020} and SCARF\cite{bahri_scarf_2022}, two leading contrastive learning solutions, and is simple and effective.
We demonstrate the utility of our pretraining strategy on the challenging task of predicting cardiac health from MR images. 
Beyond medical imaging, we show that our framework can also be applied when combining natural images and tabular data using the DVM car advertisement dataset\cite{huang_dvm-car_2021}.

Experimentally, we observe that our tool leverages morphometric features during contrastive learning. 
Morphometric features describe the size and shape of an object and therefore correlate with extractable imaging features.
We quantitatively demonstrate the importance of these features in the contrastive learning process using attribution methods, such as integrated gradients\cite{sund_intgrad}, and ablation experiments. 

Finally, we introduce a new supervised contrastive learning method called label as a feature (LaaF). 
By appending the target label as a tabular feature, our method outperforms previously published strategies that incorporate labels into the contrastive framework.
Our method is also highly flexible and can be combined with the aforementioned strategies to further improve performance.

\section{Related Work}
\label{sec:relwork}

\textbf{Self-supervised learning with images} aims to extract useful features from unlabeled data. 
Historically, this was attempted by solving hand-crafted pretext tasks such as jigsaw puzzles \cite{pang_solving_2020,taleb_multimodal_2021,zhuang_self-supervised_2019,taleb_3d_2020}, colorization \cite{larsson_learning_2017,zhang_colorful_2016,vondrick_tracking_2018}, image inpainting \cite{pathak_context_2016}, and context prediction \cite{chen_self-supervised_2019,doersch_unsupervised_2015,bai_self-supervised_2019}.
The major difficulties with using these methods is that they tend to overfit on the specifics of their pretext task, limiting their utility for downstream tasks.

\textbf{Contrastive learning} has emerged as a popular and performant successor to pretext tasks.
Contrastive learning trains encoders by generating augmented views of a sample and maximizing their projected embedding similarity while minimizing the similarity between the projected embeddings of other samples \cite{hadsell_dimensionality_2006}.
It has been popularized by implementations such as SimCLR \cite{chen_simple_2020}, MOCO \cite{he_momentum_2020}, BYOL, \cite{grill_bootstrap_2020} and others \cite{caron_unsupervised_2021,chen_exploring_2021,chen_big_2020,caron_emerging_2021,zbontar2021barlow}.
We use the contrastive framework of SimCLR as the basis for our work.

\textbf{Deep learning with tabular data} has recently begun to yield results that are competitive with classical machine learning methods \cite{borisov_deep_2022,arik2021tabnet,hollmann_tabpfn_2022}, though for many applications they still underperform simpler algorithms \cite{shwartz-ziv_tabular_2021,borisov_deep_2022}.
Self-supervised learning is being explored in the tabular domain with frameworks such as VIME\cite{yoon_vime_2020} and contrastive methods such as SubTab\cite{ucar_subtab_2021} and SCARF\cite{bahri_scarf_2022}.
We base our tabular augmentations on those used in SCARF.

\textbf{Multimodal contrastive learning with images} is becoming more important as the number of multimodal datasets increases and multimodal training strategies become more effective.
Approaches such as CLIP\cite{radford_learning_2021}, which combines images and text, are general-purpose vision models that are able to solve new tasks in a zero-shot manner.
Some of these models use internet-size datasets and are described as foundational models, such as UniCL\cite{yang_unified_2022}, Florence\cite{yuan_florence_2021}, ALIGN\cite{jia_scaling_2021}, and Wu Dao 2.0 \cite{demo_wu_nodate}.
Outside of the image-language domain, there has also been progress on multimodal contrastive learning using two different imaging modalities \cite{taleb_multimodal_2021,pielawski_comir_2020}, audio and video \cite{ma_active_2022}, video and text \cite{Zolfaghari_crossclr,xu_videoclip_2021}, and imaging and genetic data \cite{taleb_contig_2021}.
While literature on generative self-supervised tabular and imaging models \cite{antelmi2021combining}\cite{ko2022deep} exists, it is limited in scope, using only two or four clinical features.
To the best of our knowledge, there is no implementation of a contrastive self-supervised framework that incorporates both images and tabular data, which we aim to address with this work.

\textbf{Supervised learning within contrastive frameworks} has been shown to outperform the binary cross entropy loss in some cases and create more robust embeddings\cite{khosla_supervised_2020}.
Supervised contrastive learning\cite{khosla_supervised_2020} maximizes the similarity of the projected embeddings of all views in a batch from the same class.
This also addresses the problem of false negatives in contrastive learning, which is that the contrastive loss minimizes projected embedding similarity between different samples even if they are part of the same class according to a downstream task (i.e. false negatives).
By utilizing the available labels, supervised contrastive learning is able to circumvent this problem and outperforms other methods that heuristically identify and eliminate false negatives \cite{chen_incremental_2022,huynh_boosting_2022}.
We propose a solution for supervised learning in our multimodal contrastive framework that takes advantage of the unique strengths of tabular data by appending the label as a tabular feature.

\section{Methods}
\label{sec:methods}

\subsection{Contrastive Framework for Tabular and \\Imaging Data}

We base our multimodal framework on SimCLR\cite{chen_simple_2020}.
Let our dataset be $x$ and a unique sample be $j$.
Each batch contains pairs of imaging $x_{j_i}$ and tabular $x_{j_t}$ samples which are augmented.
Each augmented imaging sample $x_{j_i}$ in the batch is passed through an imaging encoder $f_{\theta_I}$ to generate the embedding $\widetilde{x_{j_i}}$.
Each augmented tabular sample $x_{j_t}$ in the batch is passed through a tabular encoder $f_{\theta_T}$ to generate the embedding $\widetilde{x_{j_t}}$.
The embeddings are propagated through separate projection heads $f_{\phi_I}$ and $f_{\phi_T}$ and brought into a shared latent space as projections $z_{j_i}$ and $z_{j_t}$ which are then L2 normalized onto a unit hypersphere.
The projections are pulled and pushed in the shared latent space according to the ``CLIP'' loss \cite{radford_learning_2021}, which maximizes the cosine similarity of projections from the same sample and minimizes the similarity of projections from different samples.
In contrast to the original InfoNCE\cite{oord_representation_2019} loss and following CLIP, we only contrast projections between modalities, never within one modality.

$i$ and $t$ can be used interchangeably and so, without loss of generality, the projection of an image is defined as 
\begin{equation}
    z_{j_i} = f_{\phi_I}(f_{\theta_I}(x_{j_i}))
\end{equation}
Considering all subjects $\NM$ in a batch, the loss for the imaging modality is
\begin{equation}  
    \ell_{i,t} = -\displaystyle\sum_{j\in\NM}log{\frac{\text{exp}(\text{cos}(z_{j_i}, z_{j_t}) / \tau)}{\displaystyle\sum_{k\in\NM,k\neq j}\text{exp}(\text{cos}(z_{j_i}, z_{k_t}) / \tau)}}.
\end{equation}
$\ell_{t,i}$ is calculated analagously and the total loss is thus
\begin{equation}
    \LM = \lambda \ell_{i,t} + (1-\lambda)\ell_{t,i}.
\end{equation}
\label{eq:combined_loss}

The images in the batch are augmented based on the standard contrastive augmentations specified in \cite{chen_simple_2020}: horizontal flips, rotations, color jitter, and resized crop. 
We do not use Gaussian blurring on the cardiac dataset in order to preserve fine-grained features in the MR images\cite{azizi_big_2021}.
To effectively augment the tabular data, a fraction of a subject's features are randomly selected to be ``corrupted'' (i.e. augmented), following \cite{bahri_scarf_2022}. 
Each corrupted feature's value is sampled with replacement from all values for that feature seen in the dataset.
Full implementation details are found in supplementary materials (SM) section \ref{app:implementation}.

\subsection{Explainability using Integrated Gradients}

To improve our understanding of the dynamics of the multimodal training, we analyze the importance of the individual tabular features in generating the embeddings.
Using test samples, we take the pretrained tabular encoder of our multimodal model and calculate the integrated gradients\cite{sund_intgrad} of each dimension of the embeddings.
This integrates the gradients of the encoder along the straightline path from a baseline sample, in our case a zero vector, to the test sample in question.
This yields the importance value of each tabular feature in generating the downstream prediction for that sample.
We then take the absolute value and calculate the mean importance of each feature across all embedding dimensions.
Categorical features have their means summed over all choices.
We use these results to categorize features and better understand how training in a multimodal setting influences unimodal performance.

\subsection{Contrastive Learning with Labels}

Incorporating labels into the contrastive learning process is typically done by modifying the loss function\cite{khosla_supervised_2020,chen_incremental_2022}.
We propose to take advantage of the unique structure of tabular data and directly append the downstream class label as a tabular feature.
We explore the benefits of combining our method with existing strategies for incorporating labels into the training process, such as supervised contrastive learning and false negative elimination.

\section{Experiments and Results}
\begin{table*}[t]
\tiny
    \caption{Performance of our framework on the tasks of myocardial infarction, coronary artery disease (CAD) and DVM car model prediction from images. Our multimodal pretrained model outperforms all other models on every task. The best performing model for every input type is displayed in \textbf{bold} font. Our method is highlighted gray.}
    \centering
    \resizebox{2\columnwidth}{!}
    {
    \renewcommand*{\arraystretch}{1.5}
    \begin{tabular}{| c | c  c | c c | c c |}
\hline
\thead{Model} & \thead{AUC (\%) \\ Frozen / Infarction} & \thead{AUC (\%)\\Trainable / Infarction} & \thead{AUC (\%)\\Frozen / CAD} & \thead{AUC (\%)\\Trainable / CAD} & \thead{Top-1 Accuracy (\%)\\Frozen / DVM} & \thead{Top-1 Accuracy (\%)\\Trainable / DVM} \\
\hline
Supervised ResNet50 & 72.37$\pm$1.80 & 72.37$\pm$1.80 & 68.84$\pm$2.54 & 68.84$\pm$2.54 & \underline{87.97$\pm$2.20} & 87.97$\pm$2.20 \\
\hline
SimCLR & \underline{73.69$\pm$0.36} & \underline{73.62$\pm$0.70} & \underline{69.86$\pm$0.21} & \underline{71.46$\pm$0.71} & 65.48$\pm$0.48 & 88.76$\pm$0.81 \\
BYOL & 69.18$\pm$0.43 & 70.69$\pm$2.09 & 66.91$\pm$0.19 & 70.66$\pm$0.22 & 59.73$\pm$0.28 & \underline{89.18$\pm$0.90} \\
SimSiam & 71.72$\pm$0.18 & 72.31$\pm$0.26 & 67.79$\pm$0.12 & 70.13$\pm$0.35 & 22.11$\pm$2.83 & 87.43$\pm$0.88 \\
BarlowTwins & 66.06$\pm$1.11 & 71.35$\pm$1.23 & 62.90$\pm$0.23 & 69.63$\pm$0.58 & 52.57$\pm$0.08 & 85.47$\pm$0.82 \\
\rowcolor{Gray}
Multimodal Imaging & \textbf{76.35$\pm$0.19} & \textbf{75.37$\pm$0.43} & \textbf{74.45$\pm$0.09} & \textbf{73.08$\pm$0.75} & \textbf{91.43$\pm$0.13} & \textbf{93.00$\pm$0.18}\\
\hline
\end{tabular}
    }
    \label{res:table_all_res}
\end{table*}

\subsection{Datasets}

As a first dataset, we used cardiac MR images and clinical information from the UK Biobank population study.
Our aim was to predict past and future risk of myocardial infarction and coronary artery disease (CAD). 
We used short axis cardiac MR images, which provide a cross-sectional view of the left and right ventricle of the heart.
The images used are two-channel 2D images whose channels are the middle baso-apical slice of the short axis cardiac MRI image at end-systolic and end-diastolic phases.
The short axis images were chosen as left ventricular function and morphometry are impacted by both CAD\cite{zabalgoitia_impact_2001} and cardiac infarction\cite{sutton_left_2000}.
Conversely, the left ventricle is a high-risk area in which early warning signs of cardiac dysfunction may be visible\cite{angeli_left_2018,tsao_left_nodate,raisi-estabragh_cardiovascular_2021}.
The images were zero-padded to 210x210 pixels and min-max normalized to a range of 0 to 1.
After augmentations (see SM section \ref{app:pretraining}), final image size was 128x128 pixels.

A subset of demographic, lifestyle and physiological features out of 5,000 data fields included in the UK Biobank dataset were selected for the tabular data. 
These features were chosen based on published correlations with cardiac outcomes. 
They include information about the subjects' diet\cite{yu_cardiovascular_2018}, physical activity\cite{mora_physical_2007}, weight\cite{powell-wiley_obesity_2021}, alcohol consumption\cite{piano_alcohols_2017}, smoking status\cite{lakier_smoking_1992}, and anxiety\cite{celano_anxiety_2016}.
Only features with at least 80\% coverage were included.
The full list of features can be found in SM section \ref{app:clinical_features}.
The continuous tabular data fields were standardized using z-score normalization with a mean value of 0 and standard deviation of 1 while categorical data was one-hot encoded.
The total size of the dataset was 40,874 unique subjects, split into 29,428 training, 7,358 validation, and 4,088 testing pairs of imaging and tabular data.

The first prediction target is myocardial infarction as defined by the International Classification of Diseases (ICD10) code.
ICD10 codes are maintained by the World Health Organization, used to record diagnoses during hospital admissions, and made available through the UK Biobank.
The second prediction target is CAD, also defined by ICD10 code.
The ICD codes used for each class are listed in SM section \ref{app:icd_codes}.
We combine past and future diagnoses since infarctions and CAD can go undiagnosed for many years and may only be recorded once a patient has to be treated for a severe cardiac event, making it difficult to establish when the disease began \cite{nesto1999screening,valensi2011prevalence}.
As both diseases are low frequency in the dataset (3\% for infarction and 6\% for CAD), finetune train splits were balanced using all positive subjects and a static set of randomly chosen negative subjects.
The test and validation sets were left untouched.

The second dataset is the Data Visual Marketing (DVM) dataset that was created from 335,562 used car advertisements\cite{huang_dvm-car_2021}.
The DVM dataset contains 1,451,784 images of cars from various angles (45 degree increments) as well as their sales and technical data.
For our task we chose to predict the car model from images and the accompanying advertisement data. 
The images were all 300x300 pixels and after augmentations (see SM section \ref{app:pretraining}) final image size was 128x128.
All fields that provided semantic information about the cars in question were included, such as width, length, height, wheelbase, price, advertisement year, miles driven, number of seats, number of doors, original price, engine size, body type, gearbox type, and fuel type.
Unique target identifying information like brand and model year were excluded.
The width, length, height and wheelbase values were randomly jittered by 50 millimeters so as not to be uniquely identifying.
We pair this tabular data with a single random image from each advertisement, yielding a dataset of 70,565 train pairs, 17,642 validation pairs, and 88,207 test pairs.
Car models with less than 100 samples were removed, resulting in 286 target classes.

To handle missing tabular data, we used an iterative multivariate imputer which models missing features as a function of existing features over multiple imputation rounds.
This was done after normalization, to ensure that the means and standard deviations were calculated only from recorded values.
The missing features were initialized with the mean and then entire columns were imputed in order from least amount of missing features, to most amount of missing features.
A regressor was fit with all other features as input and the currently examined column as dependent variable.
This process was repeated a maximum of $n$ times or until $\frac{max(abs(X_t - X_{t-1}))}{max(abs(X))} < \epsilon$, where $X_t$ is the feature vector being imputed at time point $t$ and $\epsilon$ is a provided tolerance, typically $1e^{-3}$.
Categorical data was then rounded to the nearest integer i.e. category.

\subsection{Experimental Setup}

All imaging encoders are ResNet50s\cite{he_deep_2016} that generate embeddings of size 2048.
Our multimodal model uses a tabular encoder which is a multilayer perceptron (MLP) with one hidden layer of size 2048 that generates embeddings of size 2048.
All weights are randomly initialized.
Our imaging projector is an MLP with one hidden layer of size 2048 and our tabular projector generates projections directly from the embeddings with a fully connected layer.
Projection size is 128 following \cite{chen_simple_2020}.
After pretraining, the projection head is removed and a fully connected layer to the output class nodes is added.

To evaluate the effectiveness of our model, we compare it to a fully supervised ResNet50 as well as multiple contrastive solutions such as SimCLR \cite{chen_simple_2020}, BYOL\cite{grill2020bootstrap}, SimSiam\cite{chen2021exploring}, and BarlowTwins\cite{zbontar2021barlow}.
We use linear probing of frozen networks to evaluate the quality of the learned representations\cite{chen_simple_2020,chen_incremental_2022,khosla_supervised_2020}.
We also benchmark each network while leaving all weights trainable during finetuning, as this typically improves upon the frozen counterpart and would be used in practice.
For the cardiac classification tasks we use area under the receiver operating characteristic curve (AUC) as our metric because the dataset is severely unbalanced for our targets. 
With only 3-6\% positive labels, a model that always predicts the negative class would achieve an accuracy of 94-97\%. 
For the DVM cars dataset we report top-1 accuracy as we have 280+ classes.
Results are reported as a mean and standard deviation calculated over five different seeds set during finetuning.
Both cardiac tasks were evaluated from a single pretrained model.
Full experimental details including the setup of the baseline models can be found in SM section \ref{app:implementation}. 

\begin{figure*}
  \centering
  \begin{subfigure}{0.32\linewidth}
    \includegraphics[width=\linewidth]{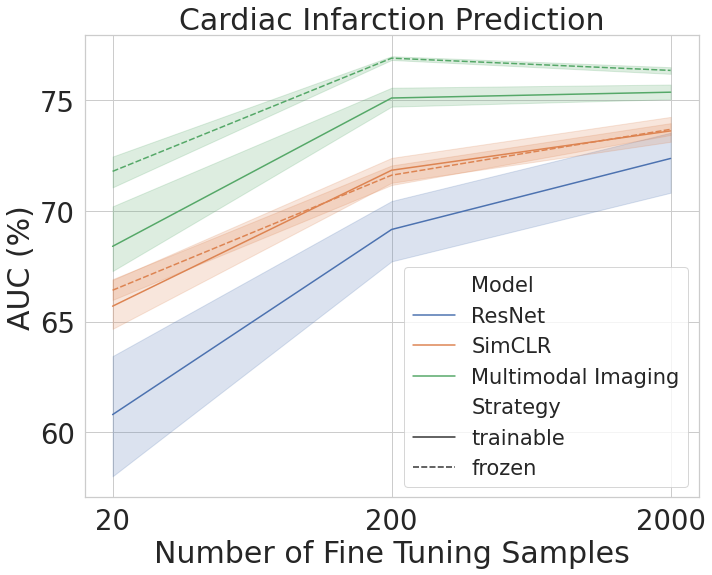}
    \label{fig:lowdata_infarction}
  \end{subfigure}
  \hfill
  \begin{subfigure}{0.32\linewidth}
    \includegraphics[width=\linewidth]{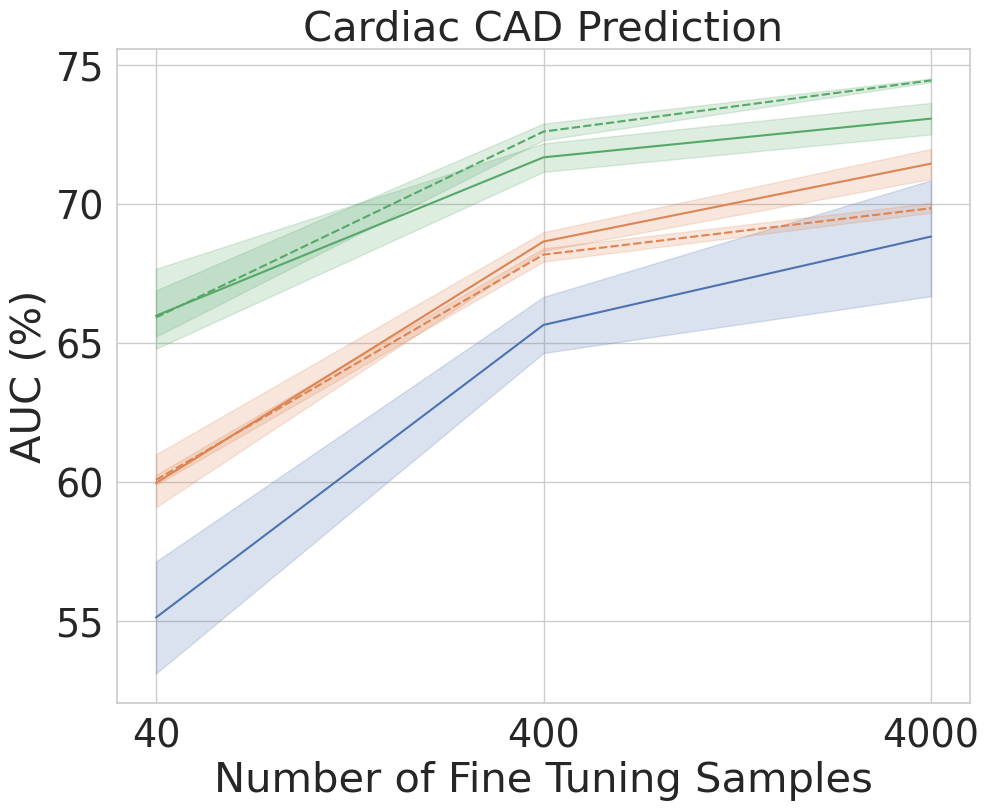}
    \label{fig:lowdata_cad}
  \end{subfigure}
  \hfill
  \begin{subfigure}{0.32\linewidth}
    \includegraphics[width=\linewidth]{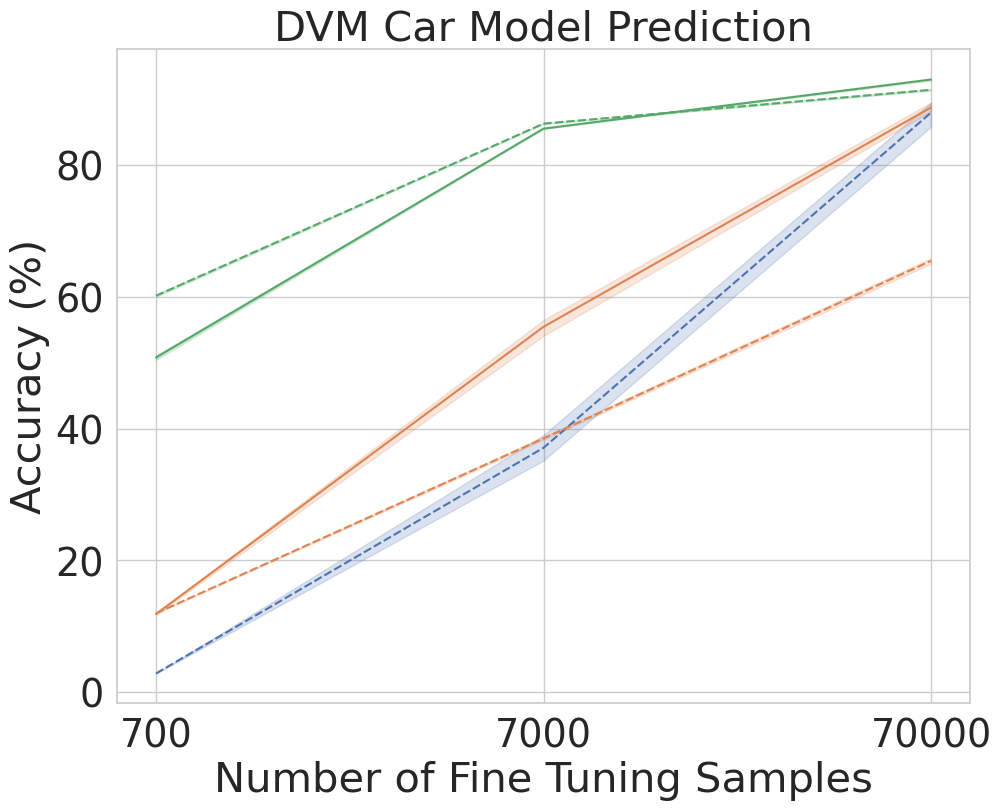}
    \label{fig:lowdata_dvm}
  \end{subfigure}
    \caption{Performance of the imaging models with different number of finetuning training samples. Shaded regions indicate 95\% confidence intervals. Pretraining with both images and tabular data excels at all data quantities and is well suited for rare disease identification when only tens or hundreds of labels are available.}
    \label{fig:low_data}
\end{figure*}

\subsection{Multimodal Pretraining Improves Unimodal Prediction}

Our main results showing the strength of our multimodal pretraining framework are found in table \ref{res:table_all_res}.
All results shown use only images as input, as this is the clinically relevant task.
Results using tabular inputs are shown in SM section \ref{app:tab_results}.
Our multimodal pretrained model substantially outperformed all other models on all three tasks and both finetuning strategies.
SimCLR generally outperformed all other contrastive strategies, highlighting our decision to base our multimodal strategy on it. 

On the cardiac tasks, the multimodal model achieved its best results when freezing the encoder.
We hypothesize that this is due to overfitting on the imaging modality during finetuning.
We suspect when provided with an imaging-only signal during finetuning that the encoder discarded features that were learned from tabular data.

When predicting the car model from images of the DVM dataset, our multimodal model outperformed other pretraining strategies by even larger margins during frozen linear probing.
This shows that for a homogeneous dataset like the DVM cars, having an additional differentiating signal such as tabular data can better align the learned features to the downstream target classes.

\begin{table*}[t]
    \caption{Frozen finetune performance of multimodal models pretrained with all features, morphometric features only, and no morphometric features. Even though the total importance of morphometric features was less than that of non-morphometric features on the cardiac task, their exclusion worsened or had equal impact on downstream performance. Best score is in \textbf{bold} font, second best \underline{underlined}.}
    \centering
    \resizebox{2\columnwidth}{!}
    {
    \renewcommand*{\arraystretch}{1.5}
    \begin{tabular}{|c|c c c c|c c c|}
    \hline
\thead{Experiment} & \thead{Tabular\\Features} & \thead{Importance\\Percentage (\%)} & \thead{AUC (\%)\\Infarction} & \thead{AUC (\%)\\CAD} & \thead{Tabular\\Features} & \thead{Importance\\Percentage (\%)} & \thead{Top-1 Accuracy (\%)\\DVM} \\
\hline
MM Imaging Baseline & 117 & 100.0 & \textbf{76.35$\pm$0.19} & \textbf{74.45$\pm$0.09} & 16 & 100.0 & \underline{91.43$\pm$0.13} \\
\hline
Morphometric Features & 24 & 47.0 & 75.22$\pm$0.30 & \underline{73.71$\pm$0.09} & 5 & 56.4 &  \textbf{92.33$\pm$0.05}\\
Non-Morphometric Features & 93 & 53.0 & \underline{75.46$\pm$0.19} & 72.18$\pm$0.25 & 11 & 43.6 & 89.14$\pm$0.24 \\
\hline
    \end{tabular}
    }
    \label{tab:dvm_phys}
\end{table*}

\subsection{Multimodal Pretraining is Beneficial in Low-Data Regimes}
\label{sec:lowdata}

When investigating rare medical conditions with very low label frequencies, models must be performant when few positive samples are available.
In order to test the performance of the learned encoders in this low-data regime, we sampled the finetuning training dataset to 10\% and 1\% of its original size, with each subset being wholly contained within each superset.
Because of the low frequencies of the positive class, this resulted in balanced training set sizes of 200 (10\%) and 20 (1\%) for myocardial infarction and 400 (10\%) and 40 (1\%) for CAD.
The test and validation set was kept identical to the full data regime.
A graphical representation of the results is shown in figure \ref{fig:low_data}.
We find that that in low-data regimes our multimodal framework generally outperforms the imaging-only contrastive method by larger margins than with full data.
This indicates improved representations that require less finetuning samples to achieve the same performance and higher utility when rare diseases are the target.
We again see that our multimodal frozen encoder consistently outperforms our trainable encoder.
We benchmark against SimCLR as it was the strongest contrastive pretraining strategy.
Comparisons against all pretraining strategies can be found in SM section \ref{app:low_data_contrastive}.

\subsection{Morphometric Features Improve Embedding Quality}

To explore why training in a multimodal fashion improves the unimodal encoders, we analyzed the contributions of the tabular features to the improved embeddings.
A unique strength of tabular data is that each of its input nodes corresponds to a single feature. 
We divided the features into two categories, morphometric and non-morphometric.
Morphometric features are related to size and shape and have direct correlates in the images, such as ventricular volume, weight or car length.
Using integrated gradients, we calculated the importance of each feature across the test samples.

\begin{figure}
    \centering
    \includegraphics[width=\linewidth]{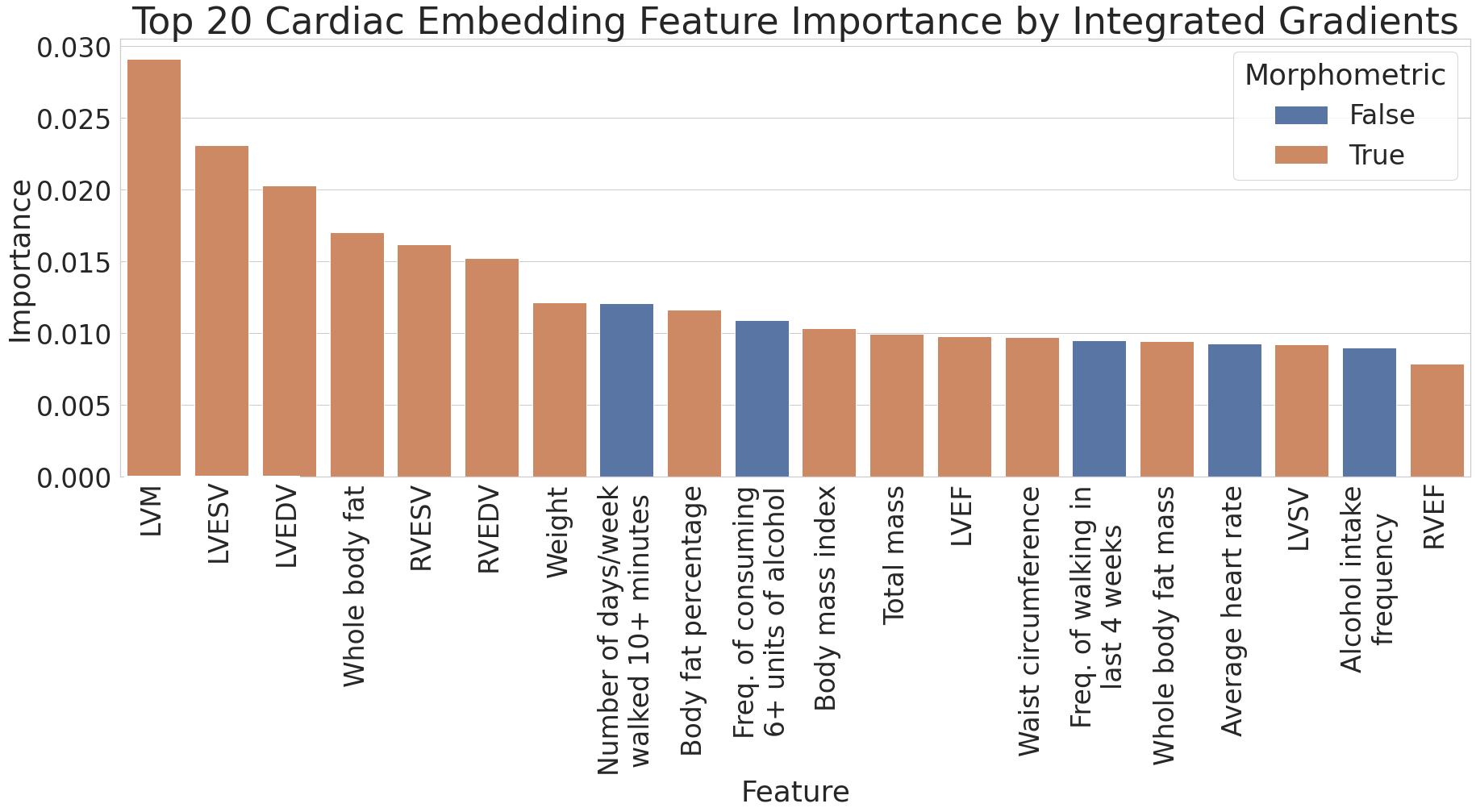}
    \caption{Top 20 most impactful features for calculating embeddings determined using integrated gradient feature attribution method. The morphometric features are colored orange and comprise 15 of the 20 most impactful features.}
    \label{fig:cardiac_ig_features}
\end{figure}

To generate the cardiac embeddings, the seven most important features were all morphometric, even though they only represent one fifth of all features.
Furthermore, all 24 morphometric features were found in the top half of the importance rankings.
These results are shown in figure \ref{fig:cardiac_ig_features} and SM section \ref{app:explain_physical_features}.

We hypothesize that the model focuses on morphometric tabular features because these have direct correlates in the images.
Extracting these features in both modalities increases the projected embedding similarity and minimizes the contrastive loss, as shown in SM section \ref{app:morph_pretrain}.

\begin{figure}
    \centering
    \includegraphics[width=\linewidth]{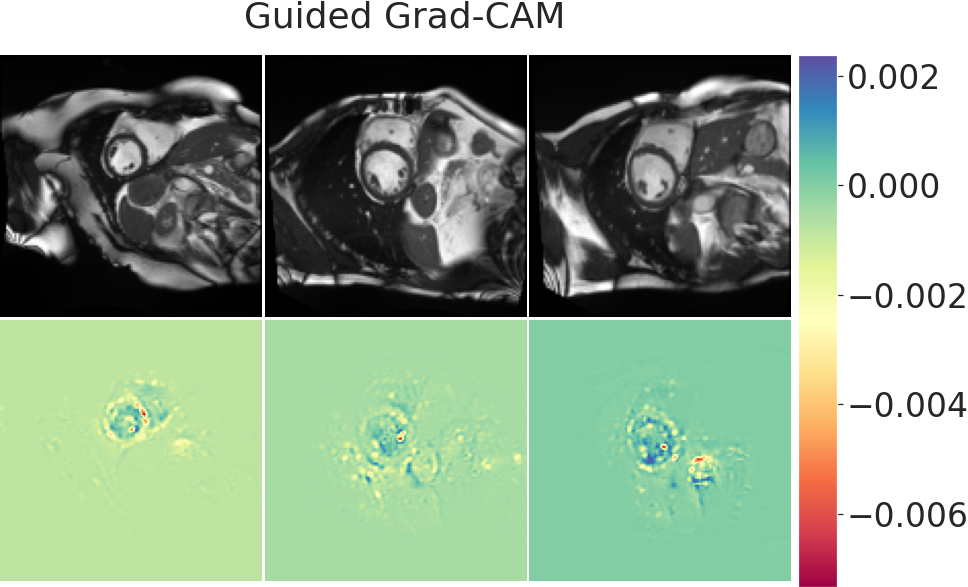}
    \caption{Guided Grad-CAM results for predicting CAD on test images. The most important features are centered around the left ventricle, matching the most important tabular features. \protect\footnotemark}
    \label{fig:gradcam}
\end{figure}

This is corroborated by the Guided GradCam results, shown in figure \ref{fig:gradcam}, where it is seen that the imaging model primarily focused on the left ventricle.
Incidentally, the three most important features according to the integrated gradients are left ventricle mass (LVM), left ventricle end systolic volume (LVESV), and left ventricle end diastolic volume (LVEDV).
To analyze the impact of these tabular features on downstream performance, we trained once with only morphometric features and once with only non-morphometric features.
We observed that the morphometric features have an outsized impact on generating the embeddings.
Table \ref{tab:dvm_phys} shows that even though they only contribute 46.99\% of the total importance and constitute 24 out of 117 features, their exclusion on CAD prediction degrades performance, and is equal to the exclusion of non-morphometric features on infarction prediction.
In general, this shows that the multimodal pretraining process is fairly robust to feature selection, especially when the total feature set is so large and there exist collinearities within the data.

\begin{figure}
    \centering
    \includegraphics[width=\linewidth]{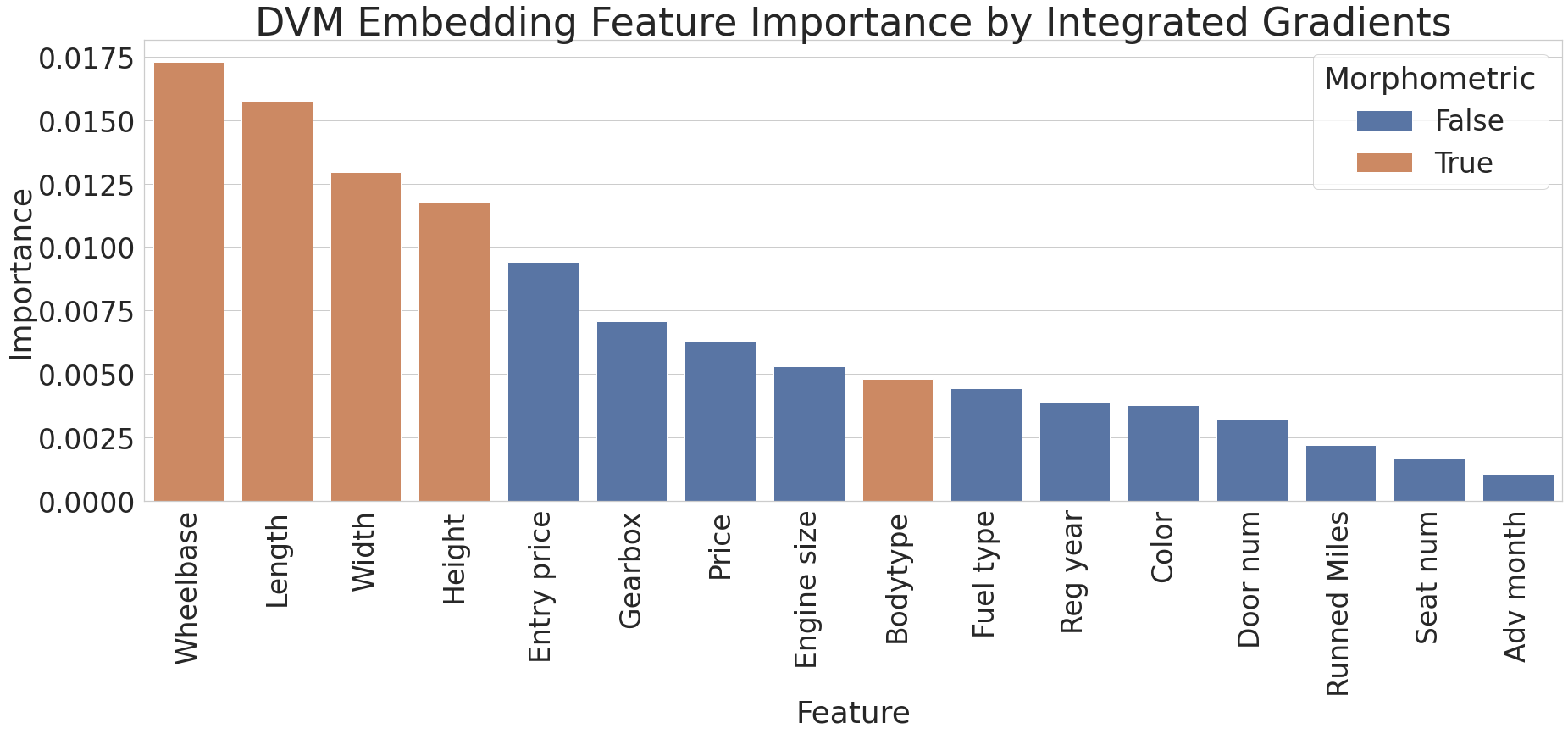}
    \caption{Impact of features for calculating DVM embeddings determined using integrated gradient feature attribution method. The morphometric features are colored orange and comprise the four most impactful features. }
    \label{fig:dvm_ig_features}
\end{figure}

Similar results are seen on the DVM dataset where the top 4 most important features are all morphometric features, despite there only being 5 morphometric features in total, as seen in figure \ref{fig:dvm_ig_features}.
Table \ref{tab:dvm_phys} shows the effect of removing these features, which led to a substantial drop in accuracy. 
When training with only morphometric features the accuracy increased highlighting their importance on tasks that are shape driven, like car model identification.
\footnotetext{These images are published with the permission of UK Biobank.}

\begin{table*}[t]
\tiny
    \caption{Frozen evaluation when incorporating labels into the contrastive pretraining process. Our Label as a Feature (Laaf) strategy consistently outperforms supervised contrastive learning (SupCon) and false negative elimination (FN Elimination), either alone or in combination. Best score is in \textbf{bold} font, second best \underline{underlined}. Our methods are highlighted gray. A dash indicates failure to converge.}
    \centering
    \resizebox{2\columnwidth}{!}
    {
    \renewcommand*{\arraystretch}{1.5}
    \begin{tabular}{|c | c | c | c c c|}
\hline
\thead{Contrastive} & \thead{Label Used} & \thead{Model} & \thead{AUC (\%)\\Infarction} & \thead{AUC (\%)\\CAD} & \thead{Top-1 Accuracy (\%)\\DVM} \\
\hline
\rowcolor{Gray}
\checkmark &  & Multimodal Imaging Baseline & \underline{76.35$\pm$0.19} & \textbf{74.45$\pm$0.09} & 91.43$\pm$0.13 \\
 & \checkmark & Supervised ResNet50 & 72.37$\pm$1.80 & 68.84$\pm$2.54 & 87.97$\pm$2.20 \\
\hline
\rowcolor{Gray}
\checkmark & \checkmark & Label as a Feature (LaaF) & \textbf{76.60$\pm$0.42} & \underline{73.76$\pm$0.31} & 93.56$\pm$0.08 \\
\checkmark & \checkmark & FN Elimination & 75.38$\pm$0.06 & 72.45$\pm$0.09 & 92.39$\pm$0.18 \\
\rowcolor{Gray}
\checkmark & \checkmark & FN Elimination + LaaF & 75.30$\pm$0.05 & 72.39$\pm$0.08 & \underline{94.07$\pm$0.05} \\
\checkmark & \checkmark & SupCon & \longdash[3] & \longdash[3] & 93.82$\pm$0.11 \\
\rowcolor{Gray}
\checkmark & \checkmark & SupCon + LaaF & \longdash[3] & \longdash[3] & \textbf{94.40$\pm$0.04} \\
\hline
    \end{tabular}
    }
    \label{tab:supcon}
\end{table*}

\subsection{Appending the Label as a Tabular Feature Boosts Supervised Contrastive Strategies}

We introduce a novel form of supervised contrastive learning by including the ground truth label as a tabular feature (LaaF).
We benchmark the effectiveness of this approach by comparing it to both supervised contrastive learning with full label information as well as false negative elimination with full label information.
The results presented in table \ref{tab:supcon} show that LaaF outperforms or rivals both strategies.

On the cardiac binary classification tasks there is a sharp class imbalance. 
97\% of the subjects are negative for myocardial infarction, leading false negative elimination to remove large portions of each batch before calculating the contrastive loss.
This leads to worse representations as batch sizes are drastically reduced during training.
As contrastive learning, and especially SimCLR, is known to be sensitive to batch sizes\cite{chen_simple_2020} this degrades downstream performance.
Supervised contrastive learning performed even worse as it did not converge during pretraining.
Again, due to the class imbalance, the supervised contrastive loss function results in a degenerate solution as approximately 97\% of the batch is projected to a single embedding.
Analogous behaviour was seen on the CAD prediction task, where 94\% of the samples are in the negative class.

On the cardiac task, LaaF performs better than false negative elimination and supervised contrastive learning, but does not offer substantial gains over the imaging baseline.
We attribute this to the fact that the cardiac setting has 120 included features, which lessens the importance of any one feature.
Additionally, imbalanced binary classification is a difficult task for supervised contrative learning as explained above.
Increasing the importance of the ground truth label in the pretraining process and adapting supervised contrastive learning to the binary case is left to future work.

On the DVM task, where we have 286 classes, the trend follows established literature.
False negative elimination improves upon the baseline and supervised contrastive learning improves upon false negative elimination\cite{chen_incremental_2022}.
Our method by itself, without modifying the loss function, surpasses false negative elimination and approaches supervised contrastive learning.
As expected, labels have a higher impact when more classes are present as shown in \cite{chen_incremental_2022}.

Importantly, adding the label as a tabular feature can also be combined with false negative elimination and supervised contrastive learning.
This highlights the flexibility of our method as it can be used with any supervised contrastive strategy.
With LaaF, we improve upon both losses and achieve our best scores on the DVM car model prediction task. 
The effect was similarly pronounced in the low data regime as shown in SM section \ref{app:laaf_low_data}.

\section{Discussion and Conclusion}

In this work we presented the first contrastive framework that combines tabular and imaging data.
Our contribution is motivated by rich clinical datasets available in biobanks that contain vast amounts of information on participants' medical history, lifestyle, and physiological measures in combination with medical images.
However, it is unfeasible to gather such detailed tabular data in a clinical setting due to time and budget constraints.
Our solution pretrains on large datasets of tabular and imaging data to boost performance during inference using only images as input.
We demonstrated the utility of our tool on the challenging task of cardiac health prediction from MR images, beating all contrastive baselines and the fully supervised baseline.
Our method also translates to the natural image domain where we showed its strength on the task of car model prediction from advertisement data. 

Through attribution and ablation experiments we showed that morphometric tabular features have outsized importance for the multimodal learning process.
We hypothesize that these features, which are related to size and shape, have direct correlates in the image and thus help minimize the multimodal self-supervised loss.
This suggests that extracting morphometric features from the images or collecting them from another source, to include them as tabular features, improves the learned representations.
Finally, we presented a simple and effective new supervised contrastive learning method when using tabular data.
Simply appending the target label as a tabular feature outperformed loss modifying strategies such as contrastive learning with false negative elimination and approached supervised contrastive learning.
This strategy can also be combined with any supervised contrastive loss modification to achieve state-of-the-art results, surpassing all other strategies.

\paragraph{Limitations}In our study we examined the benefit of our framework only for classification tasks. 
Future work should aim to test the behavior of the framework with further tasks such as segmentation and regression. 
We hypothesize that segmentation could benefit from the framework if morphometric features such as the sizes of the to-be-segmented regions are included in the tabular data and regression if morphometric features are regressed. 

A further shortcoming of this work is that we only included white subjects from the UK Biobank population dataset because other ethnicities were drastically underrepresented in the study, making up only 5\% of the total population.
Significant racial disparities in coronary infarction and CAD risk have been repeatedly found\cite{meadows_ethnic_2011,nayak_understanding_2020,ho_ethnic_2022} and could lead to spurious correlations being learned.
Future work could use balanced datasets or explore propagated biases learned with unbalanced datasets, to identify and counteract any learned biases.

\paragraph{Conclusion} In conclusion, for the first time, our work presents an effective and simple strategy to take advantage of tabular and imaging data in self-supervised contrastive learning.
Our method is particularly relevant in a clinical setting where we wish to take advantage of extensive, multimodal biobanks during pretraining and predict unimodal in practice.
We believe tabular data is an understudied and underappreciated source of data for deep learning, which is easy to collect and ubiquitous, as any numerical or categorical feature can be represented.
It is also highly interpretable due to the fact that each feature directly represents a semantic concept.
We hope that this inspires future works to unlock this untapped potential.

\subsection*{Acknowledgments}
This research has been conducted using the UK Biobank Resource under Application Number 87802.

{\small
\bibliographystyle{ieee_fullname}
\bibliography{ms}
}
\newpage\phantom{phantom}
\newpage

\input{sm}

\end{document}

%% file: sm.tex
\setcounter{section}{0}

\renewcommand{\thesection}{\Alph{section}}
\section{UK Biobank}
\label{app:ukbb}
\subsection{Clinical Cardiac Features}
\label{app:clinical_features}
The included clinical features for the cardiac prediction tasks are listed along with their UK Biobank field IDs in tables \ref{tab:tabular_features_1} and \ref{tab:tabular_features_2}. 
The features labeled \emph{extracted} were calculated using the pipeline outlined in \cite{bai2018automated, bai2018recurrent, bai2020population, petersen2017reference} with public code from \cite{bai2020population}.

\subsection{ICD Codes}
\label{app:icd_codes}
The ICD10 codes for cardiac infarction are I210, I211, I212, I213, I214, I219, and I252.

The ICD10 codes used for CAD labels are those within the categories I20-I25 - Ischemic heart diseases. 
The full list is I200, I201, I208, I209, I220, I221, I228, I229, I210, I211, I212, I213, I214, I219, I240, I248, I249, I250, I251, I252, I253, I254, I255, I256, I258, and I259.
\section{Implementation Details}
\label{app:implementation}

\subsection{Pretraining}
\label{app:pretraining}
The augmentations used during pretraining on cardiac tasks were 
\begin{itemize}
    \item RandomHorizontalFlip (probability = 0.5)
    \item RandomRotation  (degrees = 45)
    \item ColorJitter (brightness = 0.5, contrast = 0.5, saturation = 0.5)
    \item RandomResizedCrop (size = 128, scale = (0.2, 1.0))
\end{itemize}
The lower bound of the RandomResizedCrop was increased to 0.2 to ensure that a portion of the heart was in every view.

The augmentations used during pretraining on DVM car tasks were 
\begin{itemize}
    \item ColorJitter (brightness = 0.8, contrast = 0.8, saturation = 0.8, probability = 0.8)
    \item RandomGrayScale (probability = 0.2)
    \item GaussianBlur (kernel\_size = 29, sigma = (0.1, 2.0, probability = 0.5))
    \item RandomResizedCrop (size = 128, scale = (0.08, 1))
    \item RandomHorizontalFlip (probability = 0.5)
\end{itemize}

The torchvision library was used for all augmentations.
Each image was augmented during pretraining 95\% of the time.
The other 5\% of the time the image was merely resized to 128x128.
All validation and test images were simply resized to 128x128.

To effectively augment the tabular data, a fraction of a subject's features are randomly selected to be ``corrupted'' (i.e. augmented), following \cite{bahri_scarf_2022}. 
Each corrupted feature's value is sampled with replacement from all values for that feature seen in the dataset.
This is also called sampling from the empirical marginal distribution.
Categorical data was only one-hot encoded after corruption to ensure that each semantic feature had equal chance of being selected and categorical fields were not split up over multiple columns.

All contrastive pretraining models were trained for 500 epochs with a cosine annealing scheduler with warmup of 10 epochs.
The learning rate and weight decay were chosen based on validation performance with possible values being $3e^{-3}$, $3e^{-4}$, and $3e^{-5}$ for the learning rate and either $1e^{-4}$ or $1.5e^{-6}$ for the weight decay.
The Adam optimizer\cite{kingma2014adam} was used with a batch size of 512.

\begin{figure*}
  \centering
  \begin{subfigure}{0.28\linewidth}
    \includegraphics[width=\linewidth]{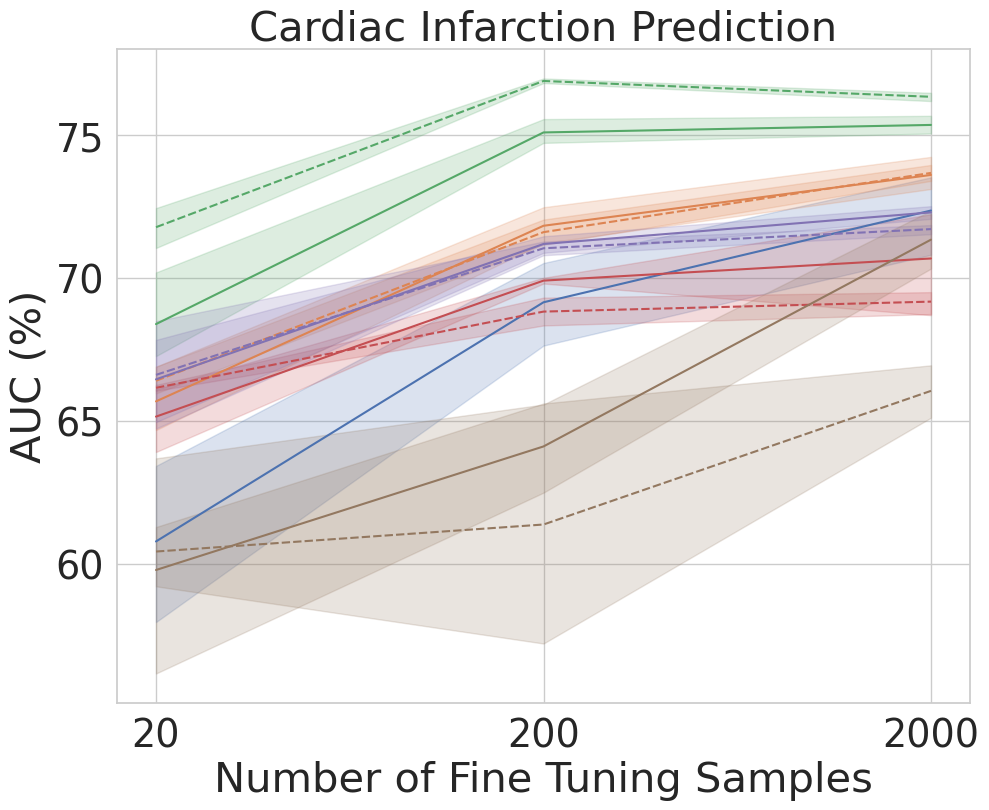}
    \label{fig:lowdata_infarction_all}
  \end{subfigure}
  \hfill
  \begin{subfigure}{0.28\linewidth}
    \includegraphics[width=\linewidth]{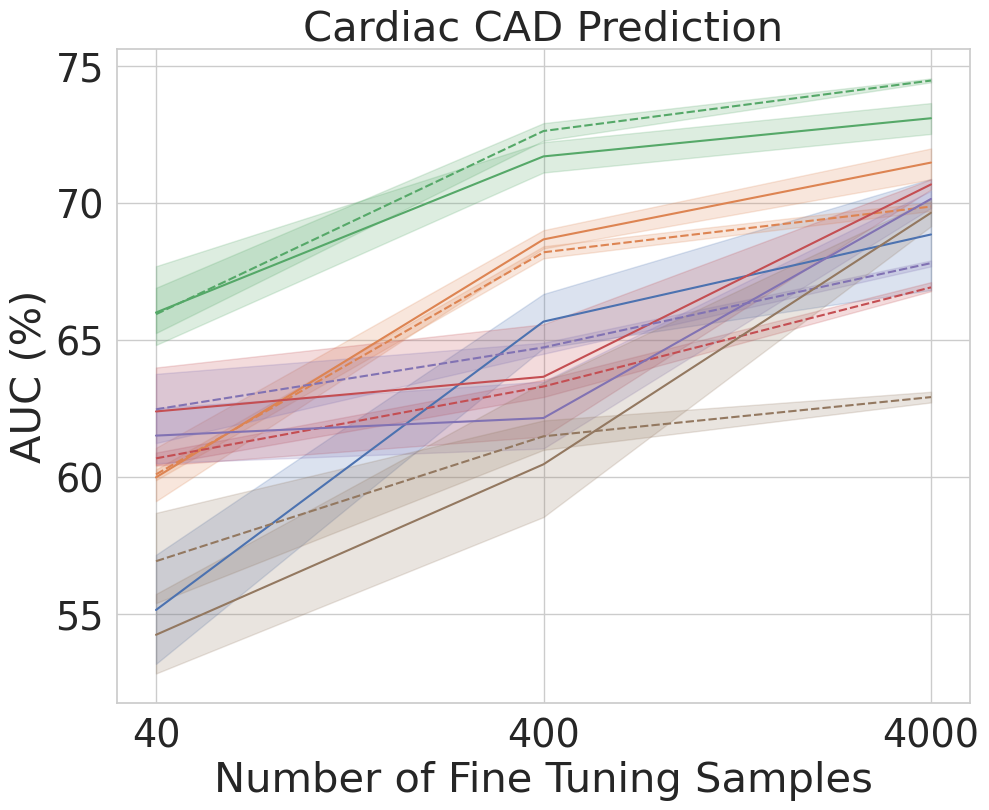}
    \label{fig:lowdata_cad_all}
  \end{subfigure}
  \hfill
  \begin{subfigure}{0.43\linewidth}
    \includegraphics[width=\linewidth]{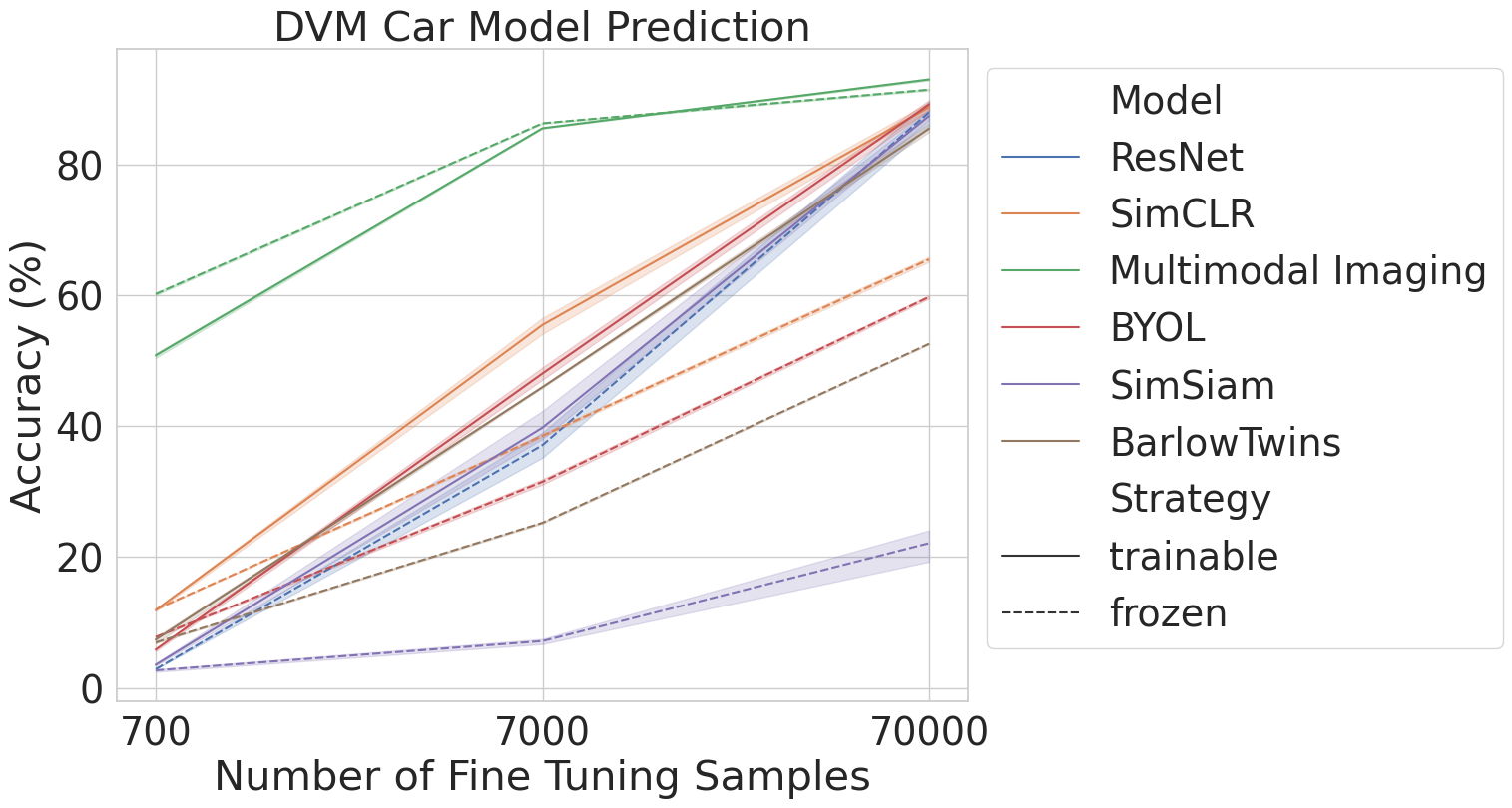}
    \label{fig:lowdata_dvm_all}
  \end{subfigure}
    \caption{Performance of the imaging models with different number of finetuning training samples. Shaded regions indicate 95\% confidence intervals. Pretraining with both images and tabular data excels at all data quantities and is well suited for rare disease identification when only tens or hundreds of labels are available.}
    \label{fig:low_data_all_models}
\end{figure*}

\paragraph{Multimodal CLIP Loss \cite{radford_learning_2021}} 
The standard CLIP loss was used as outlined in the methods section of this paper.
In our experiments, a temperature of 0.1 worked best, which follows \cite{chen_simple_2020}.
The corruption rate of the tabular augmentation was set to 0.3 after sweeping in increments of 0.1.
A learning rate of $3e^{-3}$ was used for the cardiac pretraining and $3e^{-4}$ for the DVM pretraining.
A weight decay of $1e^{-4}$ was used for the cardiac pretraining and $1.5e^{-6}$ for the DVM pretraining.

\paragraph{SimCLR \cite{chen_simple_2020}}
We used the NTXent loss which is based on the InfoNCE\cite{oord_representation_2019} loss and compares the embedding of a view against all other views in a batch.
The temperature was kept at 0.1 as outlined in the original paper.
A learning rate of $3e^{-3}$ was used for the cardiac pretraining and $3e^{-4}$ for the DVM pretraining.
A weight decay of $1e^{-4}$ was used for the cardiac pretraining and $1.5e^{-6}$ for the DVM pretraining.

\paragraph{Bootstrap Your Own Latent (BYOL) \cite{grill_bootstrap_2020}}
We used the pytorch lightning bolts implementation of BYOL.
BYOL uses an online network and a target network.
The online network has a prediction head ontop of a projection head.
The target networks weights are an exponential moving average of the online network's weights.
The loss is the cosine similarity between the prediction of the online network and the projection of the target network.
The output of the target network has a stop gradient applied so no weight updates are made outside of the exponential moving average, which is tempered by $\tau_{base}$.
The projector hidden dimension was 4096 and projector out dimension was 256.
The predictor hidden dimension was also 4096 and predictor out dimension was also 256.
$\tau_{base}$ was set to 0.9995 as we used a smaller batch size (512 vs 4096) as recommended in the original paper.
A learning rate of $3e^{-4}$ was used for the cardiac pretraining and $3e^{-4}$ for the DVM pretraining.
A weight decay of $1.5e^{-6}$ was used for all pretrainings.

\paragraph{Simple Siamese Network (SimSiam) \cite{chen_exploring_2021}}
SimSiam is similar to BYOL in that it also has a predictor on top of a projector, but it only uses a single encoder.
The prediction of the one view is compared to the projection of the other view using the cosine similarity loss with a stop gradient again being used on the side of the projection.
The projector hidden dimension was 2048 and projector out dimension was 2048.
The predictor hidden dimension was 512 and predictor out dimension was 2048, creating a bottleneck as recommended in the original paper.
A learning rate of $3e^{-4}$ was used for cardiac pretraining and $3e^{-5}$ for DVM pretraining.
A weight decay of $1.5e^{-6}$ was used for all pretrainings.

\paragraph{Barlow Twins \cite{zbontar2021barlow}}
Barlow Twins calculates the cross correlation matrix between the embeddings of the two views and pushes it towards the identity matrix, effectively maximizing similarity between views from the same subject and minimizing similarity between all other views.
The advantage of barlow twins is that it does not require a predictor head, large batches, gradient stopping or a moving average of the weights, unlike previous methods.
We use projector hidden dimensions and projector out dimensions of 8192, as recommended in the original paper.
A learning rate of $3e^{-3}$ and a weight decay of $1e^{-4}$ was used for all pretrainings.

\subsection{Finetuning}
A learning rate sweep covering 6 learning rates, ($3e^{-2}$, $1e^{-2}$, $3e^{-3}$, $1e^{-3}$, $3e^{-4}$, $1e^{-4}$) was undertaken for every model during finetuning.
The supervised models were swept in their entirety, and the contrastive models were swept during both finetuning settings, frozen and trainable.
The optimal learning rate based on validation metric performance was selected.
Early stopping based on the validation metric was used with a minimum delta of 0.0002 and a patience of 10 epochs.
The Adam optimizer without weight decay and a batch size of 512 were used.

\begin{table}[t]
    \caption{Frozen eval results when incorporating labels into the contrastive pretraining process at 100\%, 10\%, and 1\% training dataset sizes on the DVM task. Our label-as-a-feature strategy consistently outperforms supervised contrastive learning (SupCon) and false negative elimination (FN Elimination), either alone or in combination. Best score is in \textbf{bold} font, second best \underline{underlined}. Our methods are highlighted gray.}
    \centering
    \resizebox{\columnwidth}{!}
    {
    \renewcommand*{\arraystretch}{1.5}
    \begin{tabular}{| c | c c c|}
\hline
\thead{Model} & \thead{Top-1 Acc. (\%)\\DVM (100\%)} & \thead{Top-1 Acc. (\%)\\DVM (10\%)} & \thead{Top-1 Acc. (\%)\\DVM (1\%)} \\
\hline
\rowcolor{Gray}
Multimodal Baseline & 91.43$\pm$0.13 & 86.30$\pm$0.08 & 60.18$\pm$0.21 \\
Supervised ResNet50 & 87.97$\pm$2.20 & 30.69$\pm$14.02 & 2.84$\pm$0.00 \\
\hline
\rowcolor{Gray}
Label-as-a-Feature (LaaF) & 93.56$\pm$0.08 & 89.87$\pm$0.03 & \textbf{67.50$\pm$0.10} \\
FN Elim. & 92.39$\pm$0.18 & 87.61$\pm$0.07 & 63.95$\pm$0.14 \\
\rowcolor{Gray}
FN Elim. + LaaF & \underline{94.07$\pm$0.05} & \underline{89.99$\pm$0.05} & 63.37$\pm$0.70 \\
SupCon & 93.82$\pm$0.11 & 89.75$\pm$0.08 & 63.29$\pm$0.33 \\
\rowcolor{Gray}
SupCon + LaaF & \textbf{94.40$\pm$0.04} & \textbf{90.37$\pm$0.05} & \underline{64.01$\pm$0.77} \\
\hline
    \end{tabular}
    }
    \label{tab:table_low_data_laaf}
\end{table}

\begin{table*}[t]
    \caption{Performance of our framework using tabular input on the tasks of cardiac infarction, coronary artery disease (CAD), and DVM car model prediction. Our model performs similarly to SCARF on the cardiac tasks and is stronger on the DVM task. The best performing model for every input type is displayed in \textbf{bold} font and the second best is \underline{underlined}. Our method is highlighted gray.}
    \centering
    \resizebox{2\columnwidth}{!}
    {
    \renewcommand*{\arraystretch}{1.5}
    \begin{tabular}{| c | c  c | c c | c c |}
\hline
\thead{Model} & \thead{AUC (\%) \\ Frozen / Infarction} & \thead{AUC (\%)\\Trainable / Infarction} & \thead{AUC (\%)\\Frozen / CAD} & \thead{AUC (\%)\\Trainable / CAD} & \thead{Top-1 Accuracy (\%)\\Frozen / DVM} & \thead{Top-1 Accuracy (\%)\\Trainable / DVM} \\
\hline
Supervised MLP & 83.35$\pm$0.29 & 83.35$\pm$0.29 & 79.61$\pm$7.19 & 79.61$\pm$7.19 & \underline{91.99$\pm$0.10} & 91.99$\pm$0.10\\
SCARF & \underline{85.16$\pm$0.60} & \textbf{86.01$\pm$0.39} & \textbf{83.21$\pm$0.42} & \textbf{84.23$\pm$0.25} & 88.29$\pm$0.40 & \underline{92.64$\pm$0.29}\\
\rowcolor{Gray}
Multimodal Tabular & \textbf{85.76$\pm$0.27} & \underline{85.20$\pm$0.14} & \underline{83.15$\pm$0.20} & \underline{83.44$\pm$0.67} & \textbf{92.88$\pm$0.30} & \textbf{93.08$\pm$0.16} \\
\hline
\end{tabular}
    }
    \label{res:table_tab_res}
\end{table*}

\section{Low Data Contrastive Pretraining}
\label{app:low_data_contrastive}

As seen in figure \ref{fig:low_data_all_models}, our multimodal pretraining strategy excels at all data quantities and outperforms the imaging only contrastive baselines by even larger margins.
This underscores the strength of the learned representations that need minimal examples to perform on downstream classification tasks.
This makes our framework particularly well suited to rare disease classification where few positive samples are available. 

\section{Low Data Training with Label as a Feature}
\label{app:laaf_low_data}

As shown in table \ref{tab:table_low_data_laaf}, our label as a feature (LaaF) strategy for supervised contrastive learning is particularly effective in the low data regime.
LaaF consistently outperforms both supervised contrastive learning and false negative elimination, either alone or in combination with the aformentioned loss modifications.
In the very low data regime (1\% or 700 samples), LaaF by itself surpasses both SupCon and FN Elimination.

\section{Tabular Results}
\label{app:tab_results}

As shown in table \ref{res:table_tab_res}, we found that the tabular encoder trained with our multimodal framework remains competitive with SCARF.
On the cardiac tasks, when freezing the encoder, the multimodal pretrained model rivaled SCARF.
SCARF proved slightly stronger when allowing the entire network to be trainable.
On the DVM car model prediction task our multimodal framework improved upon SCARF in the frozen setting and was slightly stronger in the trainable setting.

\section{Explainability}
\label{app:explain_physical_features}
\subsection{Integrated Gradients}

\begin{figure*}
    \centering
    \includegraphics[angle=270,origin=c,width=\textwidth,height=0.68\textheight,keepaspectratio]{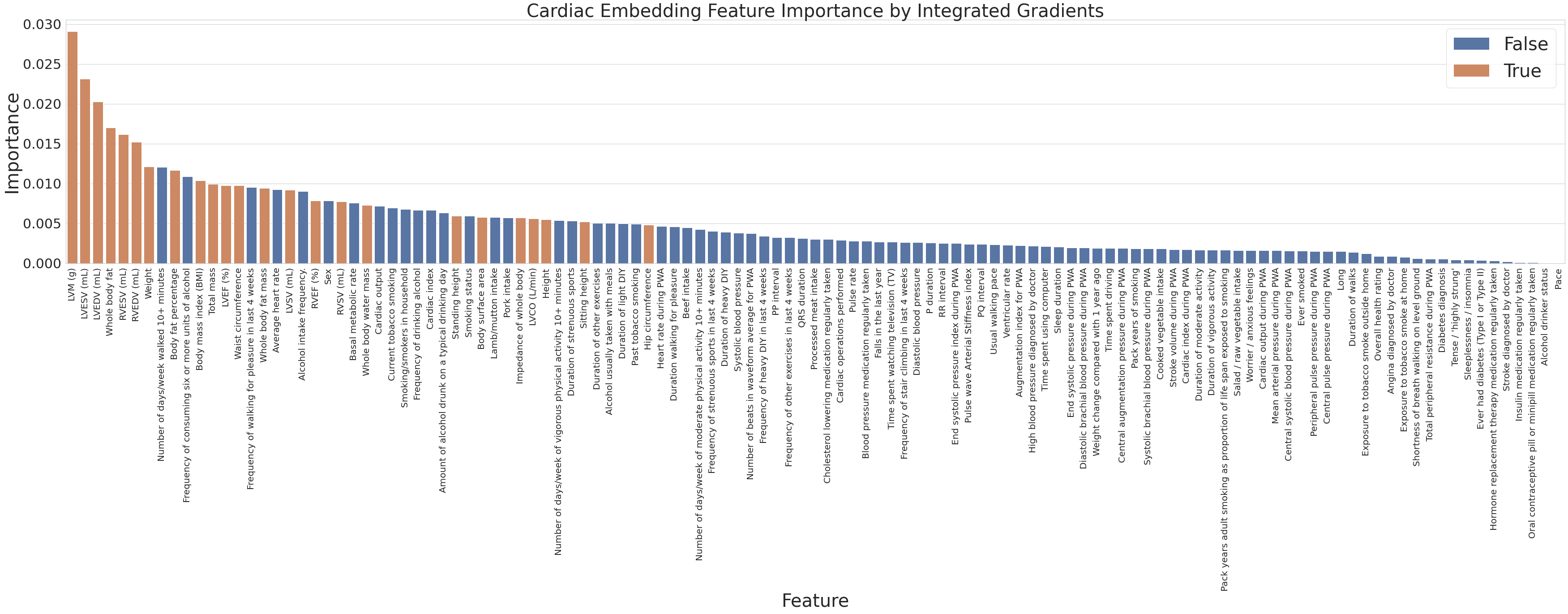}
    \caption{The integrated gradient attribution scores for tabular embeddings with respect to cardiac features. Orange color indicates morphometric feature.}
    \label{fig:ig_emb_all}
\end{figure*}

Figure \ref{fig:ig_emb_all} shows the integrated gradient attribution scores for tabular embeddings with respect to all cardiac features. 
Here the importance of the morphometric features, shown in orange, is underscored through their increased frequency towards the most important features.

\subsection{Morphometric Features Impact on Pretraining}
\label{app:morph_pretrain}
\begin{figure}
    \centering
    \includegraphics[width=\linewidth]{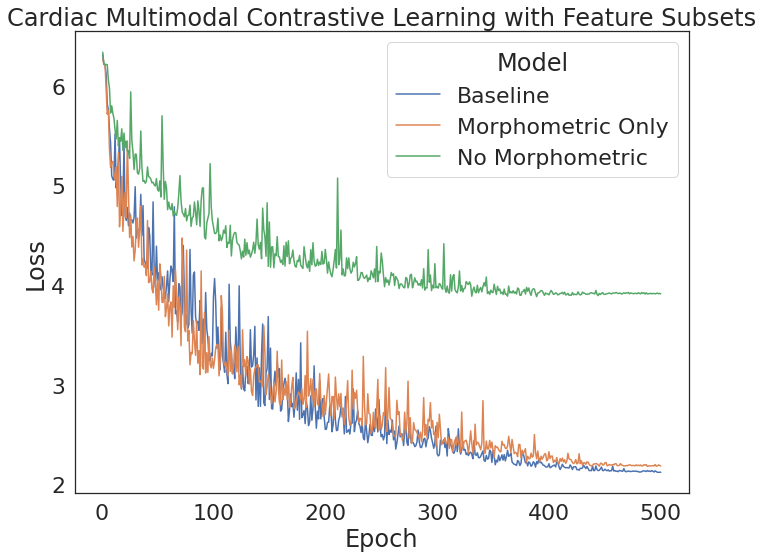}
    \caption{Contrastive loss during multimodal pretraining. Training with only morphometric features converged to a similar loss of the baseline which included all 120 features. Training with no morphometric features had markedly less similarity between the projected embeddings of the same subject, showing the importance of the morphometric features for the multimodal training process.}
    \label{fig:training_nonmorph_morph}
\end{figure}

As shown in figure \ref{fig:training_nonmorph_morph}, despite comprising less than a fourth of all features, pretraining with only morphometric features converges to almost the exact same loss as training with all 117 features.
Training without any morphometric features converged to a final loss almost twice as high.
This emphasizes the importance of morphometric features in the minimization of the CLIP loss, as they are easily extracted from images and facilitate the learning of useful features.

\newpage\phantom{test}
\newpage

\begin{table}[t]
\caption{Included cardiac features and their UK Biobank Field ID. Features labeled \emph{extracted} were calculated using the pipeline outlined in \cite{bai2018automated, bai2018recurrent, bai2020population, petersen2017reference}.}
\centering
    \resizebox{\columnwidth}{!}
    {
\begin{tabular}{| l | l |}
\hline
\thead{Tabular Feature}&\thead{UK Biobank Field ID} \\
\hline
Alcohol drinker status	&	20117  \\
Alcohol intake frequency.	&	1558  \\
Alcohol usually taken with meals	&	1618  \\
Amount of alcohol drunk on a typical drinking day	&	20403  \\
Pulse rate	&	95, 102  \\
Angina diagnosed by doctor	&	3627, 6150  \\
Augmentation index for PWA	&	12681  \\
Average heart rate	&	22426  \\
Basal metabolic rate	&	23105  \\
Beef intake	&	1369  \\
Blood pressure medication regularly taken	&	6153, 6177  \\
Body fat percentage	&	23099  \\
Body mass index (BMI)	&	23104, 21001  \\
Body surface area	&	22427  \\
Cardiac index	&	22425  \\
Cardiac index during PWA	&	12702  \\
Cardiac operations performed	&	20004  \\
Cardiac output	&	22424  \\
Cardiac output during PWA	&	12682  \\
Central augmentation pressure during PWA	&	12680  \\
Central pulse pressure during PWA	&	12678  \\
Central systolic blood pressure during PWA	&	12677  \\
Cholesterol lowering medication regularly taken	&	6177  \\
Cooked vegetable intake	&	1289  \\
Current tobacco smoking	&	1239  \\
Diabetes diagnosis	&	2443, 120007, 23104, 21001  \\
Diastolic blood pressure	&	4079, 94  \\
Diastolic brachial blood pressure during PWA	&	12675  \\
Duration of heavy DIY	&	2634  \\
Duration of light DIY	&	1021  \\
Duration of moderate activity	&	894  \\
Duration of other exercises	&	3647  \\
Duration of strenuous sports	&	1001  \\
Duration of vigorous activity	&	914  \\
Duration of walks	&	874  \\
Duration walking for pleasure	&	981  \\
End systolic pressure during PWA	&	12683  \\
End systolic pressure index during PWA	&	12684  \\
Ever had diabetes (Type I or Type II)	&	120007  \\
Ever smoked	&	20160  \\
Exposure to tobacco smoke at home	&	1269  \\
Exposure to tobacco smoke outside home	&	1279  \\
Falls in the last year	&	2296  \\
Frequency of consuming six or more units of alcohol	&	20416  \\
Frequency of drinking alcohol	&	20414  \\
Frequency of heavy DIY in last 4 weeks	&	2624  \\
Frequency of other exercises in last 4 weeks	&	3637  \\
Frequency of stair climbing in last 4 weeks	&	943  \\
Frequency of strenuous sports in last 4 weeks	&	991  \\
Frequency of walking for pleasure in last 4 weeks	&	971  \\
Heart rate during PWA	&	12673  \\
Height	&	12144  \\
High blood pressure diagnosed by doctor	&	2966, 6150  \\
Hip circumference	&	49  \\
\hline
\end{tabular}
}
    \label{tab:tabular_features_1}
\end{table}
\begin{table}[t]
\caption{Included cardiac features and their UK Biobank Field ID. Features labeled \emph{extracted} were calculated using the pipeline outlined in \cite{bai2018automated, bai2018recurrent, bai2020population, petersen2017reference}.}
\centering
    \resizebox{\columnwidth}{!}
    {
\begin{tabular}{| l | l |}
\hline
\thead{Tabular Feature}&\thead{UK Biobank Field ID} \\
\hline
Hormone replacement therapy medication regularly taken	&	6153  \\
Impedance of whole body	&	23106  \\
Insulin medication regularly taken	&	6153, 6177  \\
Lamb/mutton intake	&	1379  \\
LVCO (L/min)	&	 \emph{extracted}  \\
LVEDV (mL)	&	 \emph{extracted}  \\
LVEF ( \%)	&	 \emph{extracted}  \\
LVESV (mL)	&	 \emph{extracted}  \\
LVM (g)	&	 \emph{extracted}  \\
LVSV (mL)	&	 \emph{extracted}  \\
Mean arterial pressure during PWA	&	12687  \\
Number of beats in waveform average for PWA	&	12679  \\
Number of days/week of moderate physical activity 10+ minutes	&	884  \\
Number of days/week of vigorous physical activity 10+ minutes	&	904  \\
Number of days/week walked 10+ minutes	&	864  \\
Oral contraceptive pill or minipill medication regularly taken	&	6153  \\
Overall health rating	&	2178  \\
P duration	&	12338  \\
Pace	&	3079  \\
Pack years adult smoking as proportion of life span exposed to smoking	&	20162  \\
Pack years of smoking	&	20161  \\
Past tobacco smoking	&	1249  \\
Peripheral pulse pressure during PWA	&	12676  \\
Pork intake	&	1389  \\
PP interval	&	22334  \\
PQ interval	&	22330  \\
Processed meat intake	&	1349  \\
Pulse wave Arterial Stiffness index	&	21021  \\
QRS duration	&	12340  \\
RR interval	&	22333  \\
RVEDV (mL)	&	 \emph{extracted}  \\
RVEF ( \%)	&	 \emph{extracted}  \\
RVESV (mL)	&	 \emph{extracted}  \\
RVSV (mL)	&	 \emph{extracted}  \\
Salad / raw vegetable intake	&	1299  \\
Sex	&	31  \\
Shortness of breath walking on level ground	&	4717  \\
Sitting height	&	20015  \\
Sleep duration	&	1160  \\
Sleeplessness / insomnia	&	1200  \\
Smoking status	&	20116  \\
Smoking/smokers in household	&	1259  \\
Standing height	&	50  \\
Stroke diagnosed by doctor	&	6150  \\
Stroke volume during PWA	&	12686  \\
Systolic blood pressure	&	4080, 93  \\
Systolic brachial blood pressure during PWA	&	12674  \\
Tense / highly strung	&	1990  \\
Time spent driving	&	1090  \\
Time spent using computer	&	1080  \\
Time spent watching television (TV)	&	1070  \\
Total mass	&	23283  \\
Total peripheral resistance during PWA	&	12685  \\
Usual walking pace	&	924  \\
Ventricular rate	&	12336  \\
Waist circumference	&	48  \\
Weight	&	23098, 21002  \\
Weight change compared with 1 year ago	&	2306  \\
Whole body fat-free mass	&	23101  \\
Whole body fat mass	&	23100  \\
Whole body water mass	&	23102  \\
Worrier / anxious feelings	&	1980  \\
\hline
\end{tabular}
}
    \label{tab:tabular_features_2}
\end{table}

%% file: ms.bbl
\begin{thebibliography}{10}\itemsep=-1pt

\bibitem{alaa_cardiovascular_2019}
Ahmed~M Alaa, Thomas Bolton, Emanuele Di~Angelantonio, James~HF Rudd, and
  Mihaela Van~der Schaar.
\newblock Cardiovascular disease risk prediction using automated machine
  learning: A prospective study of 423,604 uk biobank participants.
\newblock {\em PloS one}, 14(5):e0213653, 2019.

\bibitem{angeli_left_2018}
Fabio Angeli, Paolo Verdecchia, Monica Trapasso, and Gianpaolo Reboldi.
\newblock Left ventricular hypertrophy and coronary artery calcifications: a
  dangerous duet?
\newblock {\em American Journal of Hypertension}, 31(3):287--289, 2018.

\bibitem{antelmi2021combining}
Luigi Antelmi, Nicholas Ayache, Philippe Robert, Federica Ribaldi, Valentina
  Garibotto, Giovanni Frisoni, and Marco Lorenzi.
\newblock Combining multi-task learning and multi-channel variational
  auto-encoders to exploit datasets with missing observations-application to
  multi-modal neuroimaging studies in dementia.
\newblock 2021.

\bibitem{arik2021tabnet}
Sercan~{\"O} Arik and Tomas Pfister.
\newblock Tabnet: Attentive interpretable tabular learning.
\newblock In {\em Proceedings of the AAAI Conference on Artificial
  Intelligence}, volume~35, pages 6679--6687, 2021.

\bibitem{azizi_big_2021}
Shekoofeh Azizi, Basil Mustafa, Fiona Ryan, Zachary Beaver, Jan Freyberg,
  Jonathan Deaton, Aaron Loh, Alan Karthikesalingam, Simon Kornblith, Ting
  Chen, Vivek Natarajan, and Mohammad Norouzi.
\newblock Big self-supervised models advance medical image classification.
\newblock In {\em 2021 {IEEE}/{CVF} International Conference on Computer Vision
  ({ICCV})}, pages 3458--3468. {IEEE}.

\bibitem{bahri_scarf_2022}
Dara Bahri, Heinrich Jiang, Yi Tay, and Donald Metzler.
\newblock Scarf: Self-supervised contrastive learning using random feature
  corruption.
\newblock {\em arXiv preprint arXiv:2106.15147}, 2021.

\bibitem{bai_self-supervised_2019}
Wenjia Bai, Chen Chen, Giacomo Tarroni, Jinming Duan, Florian Guitton,
  Steffen~E. Petersen, Yike Guo, Paul~M. Matthews, and Daniel Rueckert.
\newblock Self-supervised learning for cardiac {MR} image segmentation by
  anatomical position prediction.
\newblock In Dinggang Shen, Tianming Liu, Terry~M. Peters, Lawrence~H. Staib,
  Caroline Essert, Sean Zhou, Pew-Thian Yap, and Ali Khan, editors, {\em
  Medical Image Computing and Computer Assisted Intervention – {MICCAI}
  2019}, Lecture Notes in Computer Science, pages 541--549. Springer
  International Publishing.

\bibitem{bai2018automated}
Wenjia Bai, Matthew Sinclair, Giacomo Tarroni, Ozan Oktay, Martin Rajchl,
  Ghislain Vaillant, Aaron~M Lee, Nay Aung, Elena Lukaschuk, Mihir~M Sanghvi,
  et~al.
\newblock Automated cardiovascular magnetic resonance image analysis with fully
  convolutional networks.
\newblock {\em Journal of Cardiovascular Magnetic Resonance}, 20(1):1--12,
  2018.

\bibitem{bai2020population}
Wenjia Bai, Hideaki Suzuki, Jian Huang, Catherine Francis, Shuo Wang, Giacomo
  Tarroni, Florian Guitton, Nay Aung, Kenneth Fung, Steffen~E Petersen, et~al.
\newblock A population-based phenome-wide association study of cardiac and
  aortic structure and function.
\newblock {\em Nature medicine}, 26(10):1654--1662, 2020.

\bibitem{bai2018recurrent}
Wenjia Bai, Hideaki Suzuki, Chen Qin, Giacomo Tarroni, Ozan Oktay, Paul~M
  Matthews, and Daniel Rueckert.
\newblock Recurrent neural networks for aortic image sequence segmentation with
  sparse annotations.
\newblock In {\em International conference on medical image computing and
  computer-assisted intervention}, pages 586--594. Springer, 2018.

\bibitem{borisov_deep_2022}
Vadim Borisov, Tobias Leemann, Kathrin Se{\ss}ler, Johannes Haug, Martin
  Pawelczyk, and Gjergji Kasneci.
\newblock Deep neural networks and tabular data: A survey.
\newblock {\em arXiv preprint arXiv:2110.01889}, 2021.

\bibitem{caron_unsupervised_2021}
Mathilde Caron, Ishan Misra, Julien Mairal, Priya Goyal, Piotr Bojanowski, and
  Armand Joulin.
\newblock Unsupervised learning of visual features by contrasting cluster
  assignments.
\newblock {\em Advances in Neural Information Processing Systems},
  33:9912--9924, 2020.

\bibitem{caron_emerging_2021}
Mathilde Caron, Hugo Touvron, Ishan Misra, Herv{\'e} J{\'e}gou, Julien Mairal,
  Piotr Bojanowski, and Armand Joulin.
\newblock Emerging properties in self-supervised vision transformers.
\newblock In {\em Proceedings of the IEEE/CVF International Conference on
  Computer Vision}, pages 9650--9660, 2021.

\bibitem{celano_anxiety_2016}
Christopher~M Celano, Daniel~J Daunis, Hermioni~N Lokko, Kirsti~A Campbell, and
  Jeff~C Huffman.
\newblock Anxiety disorders and cardiovascular disease.
\newblock {\em Current psychiatry reports}, 18(11):1--11, 2016.

\bibitem{chen_self-supervised_2019}
Liang Chen, Paul Bentley, Kensaku Mori, Kazunari Misawa, Michitaka Fujiwara,
  and Daniel Rueckert.
\newblock Self-supervised learning for medical image analysis using image
  context restoration.
\newblock {\em Medical image analysis}, 58:101539, 2019.

\bibitem{chen_simple_2020}
Ting Chen, Simon Kornblith, Mohammad Norouzi, and Geoffrey Hinton.
\newblock A simple framework for contrastive learning of visual
  representations.
\newblock In {\em Proceedings of the 37th International Conference on Machine
  Learning}, pages 1597--1607. {PMLR}.
\newblock {ISSN}: 2640-3498.

\bibitem{chen_big_2020}
Ting Chen, Simon Kornblith, Kevin Swersky, Mohammad Norouzi, and Geoffrey~E
  Hinton.
\newblock Big self-supervised models are strong semi-supervised learners.
\newblock {\em Advances in neural information processing systems},
  33:22243--22255, 2020.

\bibitem{chen_incremental_2022}
Tsai-Shien Chen, Wei-Chih Hung, Hung-Yu Tseng, Shao-Yi Chien, and Ming-Hsuan
  Yang.
\newblock Incremental false negative detection for contrastive learning.
\newblock {\em arXiv preprint arXiv:2106.03719}, 2021.

\bibitem{chen_exploring_2021}
Xinlei Chen and Kaiming He.
\newblock Exploring simple siamese representation learning.
\newblock In {\em 2021 {IEEE}/{CVF} Conference on Computer Vision and Pattern
  Recognition ({CVPR})}, pages 15745--15753. {IEEE}.

\bibitem{chen2021exploring}
Xinlei Chen and Kaiming He.
\newblock Exploring simple siamese representation learning.
\newblock In {\em Proceedings of the IEEE/CVF conference on computer vision and
  pattern recognition}, pages 15750--15758, 2021.

\bibitem{demo_wu_nodate}
{GPT}-3 Demo.
\newblock Wu dao 2.0 {\textbar} {GPT}-3 demo.

\bibitem{doersch_unsupervised_2015}
Carl Doersch, Abhinav Gupta, and Alexei~A. Efros.
\newblock Unsupervised visual representation learning by context prediction.
\newblock In {\em 2015 {IEEE} International Conference on Computer Vision
  ({ICCV})}, pages 1422--1430. {IEEE}.

\bibitem{dugdale_time_1999}
David~C Dugdale, Ronald Epstein, and Steven~Z Pantilat.
\newblock Time and the patient--physician relationship.
\newblock {\em Journal of general internal medicine}, 14(Suppl 1):S34, 1999.

\bibitem{noauthor_german_2014}
German National Cohort (GNC)~Consortium geschaeftsstelle@ nationale-kohorte.
  de.
\newblock The german national cohort: aims, study design and organization.
\newblock {\em European journal of epidemiology}, 29(5):371--382, 2014.

\bibitem{grill2020bootstrap}
Jean-Bastien Grill, Florian Strub, Florent Altch{\'e}, Corentin Tallec, Pierre
  Richemond, Elena Buchatskaya, Carl Doersch, Bernardo Avila~Pires, Zhaohan
  Guo, Mohammad Gheshlaghi~Azar, et~al.
\newblock Bootstrap your own latent-a new approach to self-supervised learning.
\newblock {\em Advances in neural information processing systems},
  33:21271--21284, 2020.

\bibitem{grill_bootstrap_2020}
Jean-Bastien Grill, Florian Strub, Florent Altché, Corentin Tallec, Pierre~H.
  Richemond, Elena Buchatskaya, Carl Doersch, Bernardo~Avila Pires,
  Zhaohan~Daniel Guo, Mohammad~Gheshlaghi Azar, Bilal Piot, Koray Kavukcuoglu,
  Rémi Munos, and Michal Valko.
\newblock Bootstrap your own latent: A new approach to self-supervised
  learning.
\newblock ({arXiv}:2006.07733).

\bibitem{hadsell_dimensionality_2006}
R. Hadsell, S. Chopra, and Y. {LeCun}.
\newblock Dimensionality reduction by learning an invariant mapping.
\newblock In {\em 2006 {IEEE} Computer Society Conference on Computer Vision
  and Pattern Recognition - Volume 2 ({CVPR}'06)}, volume~2, pages 1735--1742.
  {IEEE}.

\bibitem{he_momentum_2020}
Kaiming He, Haoqi Fan, Yuxin Wu, Saining Xie, and Ross Girshick.
\newblock Momentum contrast for unsupervised visual representation learning.
\newblock In {\em 2020 {IEEE}/{CVF} Conference on Computer Vision and Pattern
  Recognition ({CVPR})}, pages 9726--9735. {IEEE}.

\bibitem{he_deep_2016}
Kaiming He, Xiangyu Zhang, Shaoqing Ren, and Jian Sun.
\newblock Deep residual learning for image recognition.
\newblock In {\em 2016 {IEEE} Conference on Computer Vision and Pattern
  Recognition ({CVPR})}, pages 770--778. {IEEE}.

\bibitem{ho_ethnic_2022}
Frederick~K Ho, Stuart~R Gray, Paul Welsh, Jason~MR Gill, Naveed Sattar, Jill~P
  Pell, and Carlos Celis-Morales.
\newblock Ethnic differences in cardiovascular risk: examining differential
  exposure and susceptibility to risk factors.
\newblock {\em BMC medicine}, 20(1):1--10, 2022.

\bibitem{hollmann_tabpfn_2022}
Noah Hollmann, Samuel M{\"u}ller, Katharina Eggensperger, and Frank Hutter.
\newblock Tabpfn: A transformer that solves small tabular classification
  problems in a second.
\newblock {\em arXiv preprint arXiv:2207.01848}, 2022.

\bibitem{huang_dvm-car_2021}
Jingming Huang, Bowei Chen, Lan Luo, Shigang Yue, and Iadh Ounis.
\newblock Dvm-car: A large-scale automotive dataset for visual marketing
  research and applications.
\newblock {\em arXiv preprint arXiv:2109.00881}, 2021.

\bibitem{huynh_boosting_2022}
Tri Huynh, Simon Kornblith, Matthew~R. Walter, Michael Maire, and Maryam
  Khademi.
\newblock Boosting contrastive self-supervised learning with false negative
  cancellation.
\newblock In {\em 2022 {IEEE}/{CVF} Winter Conference on Applications of
  Computer Vision ({WACV})}, pages 986--996. {IEEE}.

\bibitem{jia_scaling_2021}
Chao Jia, Yinfei Yang, Ye Xia, Yi-Ting Chen, Zarana Parekh, Hieu Pham, Quoc Le,
  Yun-Hsuan Sung, Zhen Li, and Tom Duerig.
\newblock Scaling up visual and vision-language representation learning with
  noisy text supervision.
\newblock In {\em International Conference on Machine Learning}, pages
  4904--4916. PMLR, 2021.

\bibitem{jonsson_brain_2019}
Benedikt~Atli J{\'o}nsson, Gyda Bjornsdottir, TE Thorgeirsson, Lotta~Mar{\'\i}a
  Ellingsen, G~Bragi Walters, DF Gudbjartsson, Hreinn Stefansson, Kari
  Stefansson, and MO Ulfarsson.
\newblock Brain age prediction using deep learning uncovers associated sequence
  variants.
\newblock {\em Nature communications}, 10(1):1--10, 2019.

\bibitem{khosla_supervised_2020}
Prannay Khosla, Piotr Teterwak, Chen Wang, Aaron Sarna, Yonglong Tian, Phillip
  Isola, Aaron Maschinot, Ce Liu, and Dilip Krishnan.
\newblock Supervised contrastive learning.
\newblock In H. Larochelle, M. Ranzato, R. Hadsell, M.~F. Balcan, and H. Lin,
  editors, {\em Advances in Neural Information Processing Systems}, volume~33,
  pages 18661--18673. Curran Associates, Inc.

\bibitem{kingma2014adam}
Diederik~P Kingma and Jimmy Ba.
\newblock Adam: A method for stochastic optimization.
\newblock {\em arXiv preprint arXiv:1412.6980}, 2014.

\bibitem{ko2022deep}
Wonjun Ko, Wonsik Jung, Eunjin Jeon, and Heung-Il Suk.
\newblock A deep generative--discriminative learning for multimodal
  representation in imaging genetics.
\newblock {\em IEEE Transactions on Medical Imaging}, 41(9):2348--2359, 2022.

\bibitem{lakier_smoking_1992}
Jeffrey~B Lakier.
\newblock Smoking and cardiovascular disease.
\newblock {\em The American journal of medicine}, 93(1):S8--S12, 1992.

\bibitem{larsson_learning_2017}
Gustav Larsson, Michael Maire, and Gregory Shakhnarovich.
\newblock Learning representations for automatic colorization.
\newblock In {\em European conference on computer vision}, pages 577--593.
  Springer, 2016.

\bibitem{ma_active_2022}
Shuang Ma, Zhaoyang Zeng, Daniel McDuff, and Yale Song.
\newblock Active contrastive learning of audio-visual video representations.
\newblock {\em arXiv preprint arXiv:2009.09805}, 2020.

\bibitem{meadows_ethnic_2011}
Telly~A Meadows, Deepak~L Bhatt, Christopher~P Cannon, Bernard~J Gersh, Joachim
  R{\"o}ther, Shinya Goto, Chiau-Suong Liau, Peter~WF Wilson, Genevieve
  Salette, Sidney~C Smith, et~al.
\newblock Ethnic differences in cardiovascular risks and mortality in
  atherothrombotic disease: insights from the reduction of atherothrombosis for
  continued health (reach) registry.
\newblock In {\em Mayo Clinic Proceedings}, volume~86, pages 960--967.
  Elsevier, 2011.

\bibitem{mitani_detection_2020}
Akinori Mitani, Abigail Huang, Subhashini Venugopalan, Greg~S Corrado, Lily
  Peng, Dale~R Webster, Naama Hammel, Yun Liu, and Avinash~V Varadarajan.
\newblock Detection of anaemia from retinal fundus images via deep learning.
\newblock {\em Nature Biomedical Engineering}, 4(1):18--27, 2020.

\bibitem{nayak_understanding_2020}
Aditi Nayak, Albert~J Hicks, and Alanna~A Morris.
\newblock Understanding the complexity of heart failure risk and treatment in
  black patients.
\newblock {\em Circulation: Heart Failure}, 13(8):e007264, 2020.

\bibitem{nesto1999screening}
Richard~W Nesto.
\newblock Screening for asymptomatic coronary artery disease in diabetes.
\newblock {\em Diabetes Care}, 22(9):1393, 1999.

\bibitem{mora_physical_2007}
Aaron van~den Oord, Yazhe Li, and Oriol Vinyals.
\newblock Physical activity and reduced risk of cardiovascular events.
\newblock {\em arXiv preprint arXiv:1807.03748}, 116, 2018.

\bibitem{oord_representation_2019}
Aaron van~den Oord, Yazhe Li, and Oriol Vinyals.
\newblock Representation learning with contrastive predictive coding.
\newblock {\em arXiv preprint arXiv:1807.03748}, 2018.

\bibitem{pang_solving_2020}
Kaiyue Pang, Yongxin Yang, Timothy~M. Hospedales, Tao Xiang, and Yi-Zhe Song.
\newblock Solving mixed-modal jigsaw puzzle for fine-grained sketch-based image
  retrieval.
\newblock In {\em 2020 {IEEE}/{CVF} Conference on Computer Vision and Pattern
  Recognition ({CVPR})}, pages 10344--10352. {IEEE}.

\bibitem{pathak_context_2016}
Deepak Pathak, Philipp Krahenbuhl, Jeff Donahue, Trevor Darrell, and Alexei~A.
  Efros.
\newblock Context encoders: Feature learning by inpainting.
\newblock In {\em 2016 {IEEE} Conference on Computer Vision and Pattern
  Recognition ({CVPR})}, pages 2536--2544. {IEEE}.

\bibitem{petersen2017reference}
Steffen~E Petersen, Nay Aung, Mihir~M Sanghvi, Filip Zemrak, Kenneth Fung,
  Jose~Miguel Paiva, Jane~M Francis, Mohammed~Y Khanji, Elena Lukaschuk,
  Aaron~M Lee, et~al.
\newblock Reference ranges for cardiac structure and function using
  cardiovascular magnetic resonance (cmr) in caucasians from the uk biobank
  population cohort.
\newblock {\em Journal of Cardiovascular Magnetic Resonance}, 19(1):1--19,
  2017.

\bibitem{piano_alcohols_2017}
Mariann~R Piano.
\newblock Alcohol’s effects on the cardiovascular system.
\newblock {\em Alcohol research: current reviews}, 38(2):219, 2017.

\bibitem{pielawski_comir_2020}
Nicolas Pielawski, Elisabeth Wetzer, Johan Öfverstedt, Jiahao Lu, Carolina
  Wählby, Joakim Lindblad, and Natasa Sladoje.
\newblock {CoMIR}: Contrastive multimodal image representation for
  registration.
\newblock In {\em Advances in Neural Information Processing Systems},
  volume~33, pages 18433--18444. Curran Associates, Inc.

\bibitem{powell-wiley_obesity_2021}
Tiffany~M Powell-Wiley, Paul Poirier, Lora~E Burke, Jean-Pierre Despr{\'e}s,
  Penny Gordon-Larsen, Carl~J Lavie, Scott~A Lear, Chiadi~E Ndumele, Ian~J
  Neeland, Prashanthan Sanders, et~al.
\newblock Obesity and cardiovascular disease: a scientific statement from the
  american heart association.
\newblock {\em Circulation}, 143(21):e984--e1010, 2021.

\bibitem{radford_learning_2021}
Alec Radford, Jong~Wook Kim, Chris Hallacy, Aditya Ramesh, Gabriel Goh,
  Sandhini Agarwal, Girish Sastry, Amanda Askell, Pamela Mishkin, Jack Clark,
  et~al.
\newblock Learning transferable visual models from natural language
  supervision.
\newblock pages 8748--8763, 2021.

\bibitem{raisi-estabragh_cardiovascular_2021}
Zahra Raisi-Estabragh, Nicholas~C Harvey, Stefan Neubauer, and Steffen~E
  Petersen.
\newblock Cardiovascular magnetic resonance imaging in the uk biobank: a major
  international health research resource.
\newblock {\em European Heart Journal-Cardiovascular Imaging}, 22(3):251--258,
  2021.

\bibitem{rim_deep-learning-based_2021}
Tyler~Hyungtaek Rim, Chan~Joo Lee, Yih-Chung Tham, Ning Cheung, Marco Yu,
  Geunyoung Lee, Youngnam Kim, Daniel~SW Ting, Crystal Chun~Yuen Chong,
  Yoon~Seong Choi, et~al.
\newblock Deep-learning-based cardiovascular risk stratification using coronary
  artery calcium scores predicted from retinal photographs.
\newblock {\em The Lancet Digital Health}, 3(5):e306--e316, 2021.

\bibitem{sijtsma_cohort_2022}
Salome Scholtens, Nynke Smidt, Morris~A Swertz, Stephan~JL Bakker, Aafje
  Dotinga, Judith~M Vonk, Freerk Van~Dijk, Sander~KR van Zon, Cisca Wijmenga,
  Bruce~HR Wolffenbuttel, et~al.
\newblock Cohort profile: Lifelines, a three-generation cohort study and
  biobank.
\newblock {\em International journal of epidemiology}, 44(4):1172--1180, 2015.

\bibitem{shwartz-ziv_tabular_2021}
Ravid Shwartz-Ziv and Amitai Armon.
\newblock Tabular data: Deep learning is not all you need.
\newblock {\em Information Fusion}, 81:84--90, 2022.

\bibitem{sudlow_uk_2015}
Cathie Sudlow, John Gallacher, Naomi Allen, Valerie Beral, Paul Burton, John
  Danesh, Paul Downey, Paul Elliott, Jane Green, Martin Landray, et~al.
\newblock Uk biobank: an open access resource for identifying the causes of a
  wide range of complex diseases of middle and old age.
\newblock {\em PLoS medicine}, 12(3):e1001779, 2015.

\bibitem{sund_intgrad}
Mukund Sundararajan, Ankur Taly, and Qiqi Yan.
\newblock Axiomatic attribution for deep networks.
\newblock In {\em Proceedings of the 34th International Conference on Machine
  Learning - Volume 70}, ICML'17, page 3319–3328. JMLR.org, 2017.

\bibitem{sutton_left_2000}
Martin G St~John Sutton and Norman Sharpe.
\newblock Left ventricular remodeling after myocardial infarction:
  pathophysiology and therapy.
\newblock {\em Circulation}, 101(25):2981--2988, 2000.

\bibitem{taleb_contig_2021}
Aiham Taleb, Matthias Kirchler, Remo Monti, and Christoph Lippert.
\newblock Contig: Self-supervised multimodal contrastive learning for medical
  imaging with genetics.
\newblock In {\em Proceedings of the IEEE/CVF Conference on Computer Vision and
  Pattern Recognition}, pages 20908--20921, 2022.

\bibitem{taleb_multimodal_2021}
Aiham Taleb, Christoph Lippert, Tassilo Klein, and Moin Nabi.
\newblock Multimodal self-supervised learning for medical image analysis.
\newblock In Aasa Feragen, Stefan Sommer, Julia Schnabel, and Mads Nielsen,
  editors, {\em Information Processing in Medical Imaging}, Lecture Notes in
  Computer Science, pages 661--673. Springer International Publishing.

\bibitem{taleb_3d_2020}
Aiham Taleb, Winfried Loetzsch, Noel Danz, Julius Severin, Thomas Gaertner,
  Benjamin Bergner, and Christoph Lippert.
\newblock 3d self-supervised methods for medical imaging.
\newblock In {\em Advances in Neural Information Processing Systems},
  volume~33, pages 18158--18172. Curran Associates, Inc.

\bibitem{tsao_left_nodate}
Connie~W Tsao, Philimon~N Gona, Carol~J Salton, Michael~L Chuang, Daniel Levy,
  Warren~J Manning, and Christopher~J O'Donnell.
\newblock Left ventricular structure and risk of cardiovascular events: a
  framingham heart study cardiac magnetic resonance study.
\newblock {\em Journal of the American Heart Association}, 4(9):e002188, 2015.

\bibitem{ucar_subtab_2021}
Talip Ucar, Ehsan Hajiramezanali, and Lindsay Edwards.
\newblock Subtab: Subsetting features of tabular data for self-supervised
  representation learning.
\newblock {\em Advances in Neural Information Processing Systems},
  34:18853--18865, 2021.

\bibitem{valensi2011prevalence}
Paul Valensi, Luc Lorgis, and Yves Cottin.
\newblock Prevalence, incidence, predictive factors and prognosis of silent
  myocardial infarction: a review of the literature.
\newblock {\em Archives of cardiovascular diseases}, 104(3):178--188, 2011.

\bibitem{vondrick_tracking_2018}
Carl Vondrick, Abhinav Shrivastava, Alireza Fathi, Sergio Guadarrama, and Kevin
  Murphy.
\newblock Tracking emerges by colorizing videos.
\newblock In {\em Proceedings of the European conference on computer vision
  (ECCV)}, pages 391--408, 2018.

\bibitem{xu_videoclip_2021}
Hu Xu, Gargi Ghosh, Po-Yao Huang, Dmytro Okhonko, Armen Aghajanyan, Florian
  Metze, Luke Zettlemoyer, and Christoph Feichtenhofer.
\newblock Videoclip: Contrastive pre-training for zero-shot video-text
  understanding.
\newblock {\em arXiv preprint arXiv:2109.14084}, 2021.

\bibitem{yang_unified_2022}
Jianwei Yang, Chunyuan Li, Pengchuan Zhang, Bin Xiao, Ce Liu, Lu Yuan, and
  Jianfeng Gao.
\newblock Unified contrastive learning in image-text-label space.
\newblock In {\em Proceedings of the IEEE/CVF Conference on Computer Vision and
  Pattern Recognition}, pages 19163--19173, 2022.

\bibitem{yoon_vime_2020}
Jinsung Yoon, Yao Zhang, James Jordon, and Mihaela van~der Schaar.
\newblock {VIME}: Extending the success of self- and semi-supervised learning
  to tabular domain.
\newblock In {\em Advances in Neural Information Processing Systems},
  volume~33, pages 11033--11043. Curran Associates, Inc.

\bibitem{yu_cardiovascular_2018}
Edward Yu, Vasanti~S Malik, and Frank~B Hu.
\newblock Cardiovascular disease prevention by diet modification: Jacc health
  promotion series.
\newblock {\em Journal of the American College of Cardiology}, 72(8):914--926,
  2018.

\bibitem{yuan_florence_2021}
Lu Yuan, Dongdong Chen, Yi-Ling Chen, Noel Codella, Xiyang Dai, Jianfeng Gao,
  Houdong Hu, Xuedong Huang, Boxin Li, Chunyuan Li, et~al.
\newblock Florence: A new foundation model for computer vision.
\newblock {\em arXiv preprint arXiv:2111.11432}, 2021.

\bibitem{zabalgoitia_impact_2001}
Miguel Zabalgoitia, Jens Berning, Michael~J Koren, Asbj{\o}rn St{\o}ylen,
  Markku~S Nieminen, Bj{\"o}rn Dahl{\"o}f, Richard~B Devereux, LIFE~Study
  Investigators, et~al.
\newblock Impact of coronary artery disease on left ventricular systolic
  function and geometry in hypertensive patients with left ventricular
  hypertrophy (the life study).
\newblock {\em The American journal of cardiology}, 88(6):646--650, 2001.

\bibitem{zbontar2021barlow}
Jure Zbontar, Li Jing, Ishan Misra, Yann LeCun, and St{\'e}phane Deny.
\newblock Barlow twins: Self-supervised learning via redundancy reduction.
\newblock In {\em International Conference on Machine Learning}, pages
  12310--12320. PMLR, 2021.

\bibitem{zhang_colorful_2016}
Richard Zhang, Phillip Isola, and Alexei~A Efros.
\newblock Colorful image colorization.
\newblock In {\em European conference on computer vision}, pages 649--666.
  Springer, 2016.

\bibitem{zhuang_self-supervised_2019}
Xinrui Zhuang, Yuexiang Li, Yifan Hu, Kai Ma, Yujiu Yang, and Yefeng Zheng.
\newblock Self-supervised feature learning for 3d medical images by playing a
  rubik’s cube.
\newblock In Dinggang Shen, Tianming Liu, Terry~M. Peters, Lawrence~H. Staib,
  Caroline Essert, Sean Zhou, Pew-Thian Yap, and Ali Khan, editors, {\em
  Medical Image Computing and Computer Assisted Intervention – {MICCAI}
  2019}, Lecture Notes in Computer Science, pages 420--428. Springer
  International Publishing.

\bibitem{Zolfaghari_crossclr}
Mohammadreza Zolfaghari, Yi Zhu, Peter Gehler, and Thomas Brox.
\newblock Crossclr: Cross-modal contrastive learning for multi-modal video
  representations.
\newblock In {\em Proceedings of the IEEE/CVF International Conference on
  Computer Vision}, pages 1450--1459, 2021.

\end{thebibliography}
